\documentclass[journal,twoside,web]{ieeecolor}
\definecolor{subsectioncolor}{rgb}{0,0.541,0.855}
\makeatletter
\newcommand\arxiv@hd[3]{\rlap{\raisebox{#1}[0pt][0pt]{\color{subsectioncolor}\rule{43pc}{1pt}}}\rlap{\raisebox{#2}[0pt][0pt]{\hb@xt@43pc{\scriptsize #3}}}\hfill}
\def\ps@titlepagestyle{\def\@oddfoot{}\def\@evenfoot{}%
  \def\@oddhead{\hspace*{7pt}\arxiv@hd{9.5pt}{14.25pt}{\hfill\textsf{\thepage}}}%
  \let\@evenhead\@oddhead}
\def\ps@headings{\def\@oddfoot{}\def\@evenfoot{}%
  \def\@oddhead{\arxiv@hd{9.75pt}{2.59pt}{\hfill\textsf{\thepage}}}%
  \def\@evenhead{\arxiv@hd{9.75pt}{2.59pt}{\textsf{\thepage}\hfill}}}
\makeatother
\usepackage{cite}
\usepackage{amsmath,amssymb,amsfonts}
\usepackage{textcomp}
\usepackage{multirow}%
\usepackage{mathrsfs}%
\usepackage{textcomp}%
\usepackage{manyfoot}%
\usepackage{booktabs}%
\usepackage{algorithm}%
\usepackage{algorithmicx}%
\usepackage{algpseudocode}%
\usepackage{listings}%
\usepackage[utf8]{inputenc}
\usepackage[T1]{fontenc}
\usepackage{lineno}
\usepackage{subcaption}
\usepackage{graphicx}
\usepackage{url}
\usepackage{ulem}
\def\BibTeX{{\rm B\kern-.05em{\sc i\kern-.025em b}\kern-.08em
    T\kern-.1667em\lower.7ex\hbox{E}\kern-.125emX}}
\begin{document}
\title{SegSTRONG-C: Segmenting Surgical Tools Robustly on Non-Adversarial Generated ``Corruptions''---An EndoVis'24 Challenge}
\author{Hao Ding*, Yuqian Zhang*, Tuxun Lu, Ruixing Liang, Hongchao Shu, Lalithkumar Seenivasan, Yonghao Long, Qi Dou, Cong Gao, Yicheng Leng, Seok Bong Yoo, Eung-Joo Lee, Zijian Wu, Yuxin Chen, Septimiu E. Salcudean, Samra Irshad, Shadi Albarqouni, Seong Tae Kim, Yueyi Sun, An Wang, Long Bai, Hongliang Ren, Ihsan Ullah, Ho-Gun Ha, Attaullah Khan, Hyunki Lee, Satoshi Kondo, Satoshi Kasai, Kousuke Hirasawa, Sita Tailor, Ricardo Sanchez-Matilla, Imanol Luengo, Tianhao Fu, Jun Ma, Bo Wang, Marcos Fernández-Rodríguez, Estevao Lima, João L. Vilaça, Negin Ghamsarian, Klaus Schoeffmann, Raphael Sznitman, Mathias Unberath
\thanks{*Hao Ding and Yuqian Zhang contributed equally to this work. Corresponding author: Hao Ding (e-mail: hding15@jhu.edu).}
\thanks{\textbf{Organizers:} Hao Ding, Yuqian Zhang, Tuxun Lu, Ruixing Liang, Hongchao Shu, Lalithkumar Seenivasan, and Mathias Unberath were with Johns Hopkins University, Baltimore, USA; Yonghao Long and Qi Dou were with The Chinese University of Hong Kong, Hong Kong, China; Cong Gao was with Intuitive Surgical Inc., Sunnyvale, USA. This challenge is supported by the Collaborative Research Agreement with the Multi-Scale Medical Robotics Center, The Chinese University of Hong Kong, Intuitive Surgical Inc., and Johns Hopkins University Internal Funds.}
\thanks{\textbf{VSI:} Yicheng Leng and Eung-Joo Lee were with The University of Arizona, Tucson, USA; Seok Bong Yoo was with Chonnam National University, Gwangju, Republic of Korea.}
\thanks{\textbf{UBCRCL:} Zijian Wu, Yuxin Chen, and Septimiu E. Salcudean were with The University of British Columbia, Vancouver, Canada.}
\thanks{\textbf{SAM\_KHU:} Samra Irshad and Seong Tae Kim were with Kyung Hee University, Yongin-si, Republic of Korea; Shadi Albarqouni was with University Hospital Bonn, Bonn, Germany, and Helmholtz AI, Munich, Germany.}
\thanks{\textbf{Medical\_Mechatronics:} Yueyi Sun was with Beijing Institute of Technology, Beijing, China; An Wang was with Shenzhen Research Institute of CUHK, Shenzhen, China, and The Chinese University of Hong Kong, Hong Kong, China; Long Bai and Hongliang Ren were with The Chinese University of Hong Kong, Hong Kong, China.}
\thanks{\textbf{CCG\_DGIST:} Ihsan Ullah, Ho-Gun Ha, Attaullah Khan, and Hyunki Lee were with Daegu Gyeongbuk Institute of Science and Technology, Daegu, Republic of Korea.}
\thanks{\textbf{SK:} Satoshi Kondo was with Muroran Institute of Technology, Hokkaido, Japan; Satoshi Kasai was with Fujita Health University, Aichi, Japan; Kousuke Hirasawa was with Konica Minolta, Inc., Tokyo, Japan.}
\thanks{\textbf{Tailor:} Sita Tailor was with University College London, London, U.K.; Ricardo Sanchez-Matilla and Imanol Luengo were with Medtronic, London, U.K.}
\thanks{\textbf{FightTumor:} Tianhao Fu was with Villanova College, King City, Canada, Project Neura, Toronto, Canada, and the Vector Institute, Toronto, Canada; Jun Ma and Bo Wang were with Project Neura, Toronto, Canada, the University of Toronto, Toronto, Canada, and the Vector Institute, Toronto, Canada.}
\thanks{\textbf{ICVS-2AI:} Marcos Fernández-Rodríguez was with the University of Minho, Braga, Portugal, PT Government Associate Laboratory, Braga/Guimarães, Portugal, and the Polytechnic Institute of Cávado and Ave, Barcelos, Portugal; Estevao Lima was with the University of Minho, Braga, Portugal, and PT Government Associate Laboratory, Braga/Guimarães, Portugal; João L. Vilaça was with the Polytechnic Institute of Cávado and Ave, Barcelos, Portugal, and Intelligent Systems Associate Laboratory, Guimarães, Portugal.}
\thanks{\textbf{AIMI:} Negin Ghamsarian and Raphael Sznitman were with the University of Bern, Bern, Switzerland; Klaus Schoeffmann was with Klagenfurt University, Klagenfurt, Austria.}
}
\maketitle

\begin{abstract}
Surgical data science has seen rapid advancement with the excellent performance of end-to-end deep neural networks (DNNs). Despite their successes, DNNs have been proven susceptible to minor ``corruptions,'' substantially impairing the model's performance, introducing a major concern for the translation of cutting-edge technology, especially in high-stakes scenarios. 
We introduce the SegSTRONG-C challenge dedicated to better understanding model deterioration under unforeseen but plausible non-adversarial ``corruption'' and the capabilities of contemporary methods that seek to improve it. Built on a dataset generated through counterfactual robotic replay, SegSTRONG-C provides paired clean and ``corrupted'' samples, enabling reproducible evaluation of model robustness. Participants are challenged to train tool segmentation algorithms on ``uncorrupted'' data and evaluate them on ``corrupted'' test domains for the binary robot tool segmentation task.
Through comprehensive baseline experiments and participating submissions from widespread community engagement, SegSTRONG-C reveals key themes for model failure and identifies promising directions for improving robustness. The challenge winners achieved an average of 0.9394 DSC and 0.9301 NSD across the unreleased test sets with ``corruption'' types: bleeding, smoke, and low brightness. This highlights how prior knowledge, customized training strategies, and architectural choice can be leveraged to improve robustness. 
In conclusion, the SegSTRONG-C challenge has identified practical approaches for enhancing model robustness. However, most approaches rely on conventional techniques that have known limitations. Looking ahead, we advocate for expanding intellectual diversity and creativity in non-adversarial robustness beyond data augmentation, calling for new paradigms that enhance universal robustness to unforeseen ``corruptions'' to facilitate richer applications in surgical data science.
\end{abstract}


\section{Introduction}\label{sec:intro}
Surgical data science is now a well-established subfield of AI-assisted medicine and has enjoyed increasing popularity and opportunities through the rapid advancement of end-to-end deep neural networks (DNNs) for image and video processing. Tool segmentation (robotic or otherwise) in surgical videos is among the most fundamental surgical data science tasks because it plays an enabling role for numerous downstream analyses and applications, such as gesture and phase recognition~\cite{liu2025deep,van2021gesture}, skill assessment~\cite{lam2022machine}, mixed reality feedback~\cite{magalhaes2024mixed}, and surgical automation~\cite{schmidgall2025will}, among others. While different downstream applications require different levels of accuracy in segmentation~\cite{ghanekar2025video,park2024towards}, improving the overall task performance and preventing model failure has been the focus of the research community. DNNs~\cite{mo2022review} have dominated semantic segmentation ever since the emergence of deep learning, fueled in large parts by the introduction of public and well-annotated datasets of surgical scenes together with the associated community grand challenges, many of which are under the EndoVis umbrella~\cite{allan20202018, Allan2019ENDOVIS17}. These challenges have established that, generally, methods that achieve state-of-the-art performance on general computer vision benchmarks~\cite{everingham2010pascal,zhou2019semantic} also exhibit excellent performance on surgical video. However, all of these previous results make the strong assumption that training and test data are highly similar and do not contain any (or at least any considerable) ``corruption'' to the image. This means that current challenges cannot assess the non-adversarial robustness of contemporary tool segmentation algorithms to various image ``corruption,'' such as tool occlusion due to smoke or bleeding.
This is a severe shortcoming of current benchmarks because in surgical video, these ``corruptions''\footnote{It is worth mentioning that while the community has been using the word ``corruption'' to describe this type of degradation caused by adverse environmental conditions~\cite{dong2023benchmarking,li2022all,rusak2020simple,schneider2020improving}, we argue that it is not rigorous use of the word as it should indicate that there is an error in the image content and that it does not correctly represent the physical world being sensed by the camera. To maintain a consistent use of the word while maintaining academic rigor, we add quotation marks for all words related to ``corruption'' in this paper.} are not the exception -- they are the norm. Because it is now well established that DNNs are highly susceptible to even minor ``corruptions'' resulting in considerably impaired performance~\cite{drenkow2021systematic}, it is critical to better understand this vulnerability and the approaches one may take to enhance model robustness.

While adversarial~\cite{malik2024systematic} and non-adversarial robustness~\cite{drenkow2021systematic,gojic2023non} have raised researchers' attention for years in general computer vision research, especially for autonomous driving~\cite{vargas2021overview,gella2024weatherproof,gupta2024robust,diaz2022ithaca365,sakaridis2021acdc,zhang2025comprehensively}, due to the lack of evaluation benchmarks, the best practice for developing non-adversarially robust algorithms for surgical analysis in general and robot tool segmentation in particular remains unestablished. The current datasets and data collection/synthesis strategies used to evaluate non-adversarial robustness are insufficient. The most common approaches, chief among them ImageNet-C~\cite{hendrycks2019benchmarking}, rely on synthetically crafted ``corruptions'' based on common augmentation techniques (such as noise injection and contrast variations)~\cite{drenkow2021systematic}. These methods do not accurately reflect the interactions of various ``corruption'' factors during image formation (such as reduced brightness not only affecting intensity but also noise due to higher sensor sensitivity)~\cite{drenkow2024causality,drenkow2025detecting} and do not offer sufficiently realistic simulation of the ``corruptions'' most relevant to robotic surgery and surgical data science. The latest Cholectrack20~\cite{nwoye2025cholectrack20} dataset contains some natural ``corruption'' types like smoke and bleeding that happen in real surgery, providing possible evaluation cases. However, these ``corruptions'' and tool poses are also confounded by the ongoing surgical actions, e.g., smoke is only generated by the interaction of the tissue and energized tools, where the tools' locations are usually close to the target tissues. An assessment based on these cases may be subject to confounding bias, resulting in a less rigorous assessment.
In this work, we present the SegSTRONG-C benchmark to draw research attention to the important issue of increasing non-adversarial robustness in surgical video analysis and to evaluate the robustness of contemporary and new approaches under non-adversarial ``corruptions'' that commonly happen in the surgical scene. This approach contributes to a more trustworthy assessment and analysis of where surgical tool segmentation algorithms stand when dealing with potential ``corruptions'' of the image. 

The SegSTRONG-C challenge introduces a dataset that provides robotic surgical scenes with and without ``corruptions'' that are created in a counterfactual setting by leveraging the automation capabilities afforded by robotic systems. Specifically, we create a dataset comprising paired surgical video sequences under clean and ``corrupted'' conditions, which is created by replaying the same robotic actions under a multitude of ``corruption'' conditions: 1) ``Uncorrupted'' video sequences, 2) variation in background, 3) variation in brightness, 4) presence/absence of blood, and 5) presence/absence of smoke. In contrast to all prior work, these ``corruptions'' are not synthetically simulated but introduced in the real world, resulting in individual ``corruptions'' potentially affecting multiple different image properties (e.g., reduced brightness potentially affecting noise, as in the example above). This dataset builds the foundation for the SegSTRONG-C challenge, where participants are challenged to develop algorithms solely on ``uncorrupted'' sequences but are evaluated and ranked based on their performance on ``corrupted'' videos for the robot tool segmentation task. 

We conducted baseline experiments with various representative DNNs, including end-to-end trained convolutional neural networks (CNNs), transformers, and pre-trained foundation models. In the study, we find that non-adversarial ``corruptions'' lead to noticeable performance degradation in end-to-end DNNs. We conduct an additional study that adds training data from the ``corrupted'' domain to explore the reason for model degradation and a potential mitigation strategy. We summarize the main findings here and provide a thorough quantitative and qualitative analysis in Section~\ref{sec:results_baseline}:
\begin{itemize}
    \item Non-adversarial ``corruptions'' lead to substantial performance degradation in end-to-end DNNs and the foundation models.
    \item Domain shift and reduced visibility make major contributions to the performance degradation.
\end{itemize}

Our challenge received high community participation in a 4-month running window with 40 registrations and 10 final submissions. According to the challenge results, we summarize the following two main aspects inspired by our observations across all submissions and quantitative and qualitative results in Section~\ref{sec:results_submission}: Data augmentation and architecture enhancement.

To consolidate the inspiration from baseline analysis and challenge results into concrete insight for developing robust algorithms in surgical applications, we design and conduct precisely controlled experiments in data augmentation and architecture design with the representative architectures we selected from the baseline analysis. We summarize the main findings here and provide a thorough quantitative and qualitative analysis in Section~\ref{sec:results_validation}:
\begin{itemize}
    \item General data augmentation provides foundational robustness and should be applied in the training of all models. 
    \item Customized data augmentation not only provides targeted robustness for specific non-adversarial ``corruptions'' but also improves overall robustness.
    \item Simply improving model capacity does not improve the model's robustness.
    \item Pretraining facilitates model optimization and more robust feature extraction. 
\end{itemize}

In summary, the SegSTRONG-C challenge, a benchmark targeted at a comprehensive evaluation of the non-adversarial robustness of contemporary tool segmentation models, confirmed that non-adversarial robustness is an important concern for vision-based surgical data science. The severe performance deterioration of the baseline models emphasizes the need for more targeted development efforts to increase robustness, or at least, dedicated evaluations to assess non-adversarial robustness, especially in more translational research. The challenge also identified several strategies that are reasonably practical to enhance model robustness. They include model selection and training augmentations. Augmentations especially offer utility if the ``corruption'' type is well-known and can be simulated with reasonable fidelity as part of data augmentation.  However, breaking the assumption that the ``corruption'' type is known during model development presents a more complex challenge, the solution to which will require a more universal level of non-adversarial robustness. Approaches to achieve this level of robustness are being attempted, e.g., through the vision foundation models that are trained with large-scale data or the incorporation of geometric constraints or other forms of prior knowledge. Overall, we argue for increasing the intellectual diversity and creativity of new approaches to address non-adversarial robustness beyond data augmentation or training set scale. Ultimately, we believe that these new paradigms will not only provide enhanced and more universal robustness to ``corruptions'' of any kind but, through their more complete modeling (e.g., by incorporating robot kinematics or other multi-modal data sources), enable richer downstream applications to benefit surgical data science as a whole. 

\section{Method}
\subsection{Dataset}

\begin{figure}[ht]
\centering
\includegraphics[width=\linewidth]{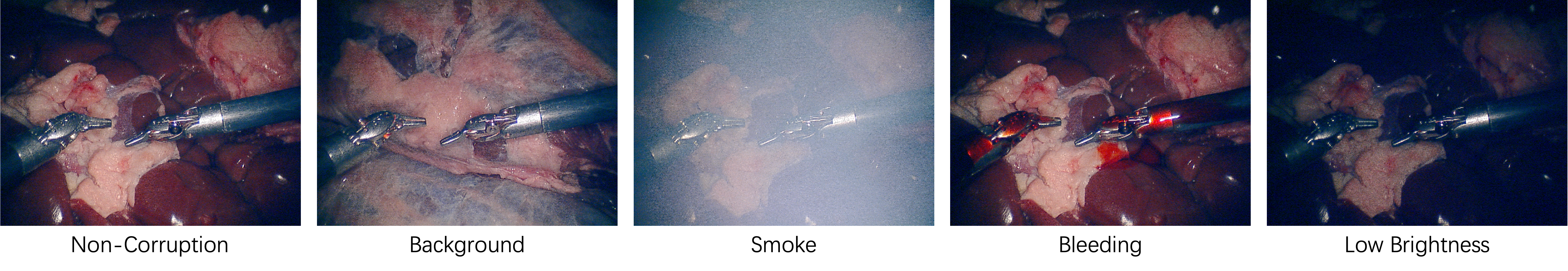}
\caption{Example images for ``uncorrupted'' image and non-adversarial ``corruptions''(background, smoke, bleeding, and low brightness).}
\label{fig:eye_candy}
\end{figure}

The dataset consists of mock endoscopic video sequences with corresponding binary segmentation masks for the robot tool segmentation task. We mock the endoscopic scene with two patient-side manipulators (PSMs) from the da Vinci surgical robot and animal tissue backgrounds to ensure a photo-realistic appearance. We manually teleoperate the robot to generate the trajectory in free space. The binary segmentation masks serve as ground truth annotations for surgical tools. The dataset consists of $17$ sequences, collected from different robot and camera configurations and different robot trajectories. Each sequence is collected at $10$ frames per second and consists of $300$ ``uncorrupted'' frames for the left camera and $300$ ``uncorrupted'' frames for the right camera. We collect all ``corrupted'' versions (background, smoke, bleeding, and low brightness) for each sequence under the same configuration and trajectory. We provide $11$ sequences of the dataset with only ``uncorrupted'' frames as the train set. We provide $3$ video sequences with ``uncorrupted'' frames and corresponding frames with background ``corruption'' as the validation set. We use the final 3 sequences with certain ``corruptions''(smoke, bleeding, and low brightness) as the test set. The models submitted by the challenge participants are tested on this test set consisting of sequences with photo-realistic non-adversarial ``corruptions.'' The models are tested on each corruption separately. Table~\ref{tab:dataset} shows the summary of the dataset. All ``corrupted'' versions for the training and validation are released and can be found on the challenge website \url{segstrongc.cs.jhu.edu}. The example images are shown in Figure~\ref{fig:eye_candy}.

\begin{table*}[ht]
  \centering
  \caption{Dataset summary. "300 + 300" means 300 frames for the left camera and 300 frames for the right camera. "-" means not provided during the challenge.}
  \resizebox{\textwidth}{!}{
    \begin{tabular}{c|c|c|c|c|c|c|c}
      \toprule
       \multirow{2}{*}{\textbf{Split}} & \textbf{Camera/robot} & \textbf{Trajectory} & \textbf{Non-} &\multirow{2}{*}{\textbf{Background}} & \multirow{2}{*}{\textbf{Smoke}} & \multirow{2}{*}{\textbf{Bleeding}} & \textbf{Low} \\ & \textbf{configuration id} & \textbf{id} & \textbf{``corruption''} & &  &  & \textbf{brightness} \\
      \midrule
      \multirow{11}{*}{\textbf{Train set}}
      & 3 & 1 & 300 + 300 & - & - & - & -\\
      & 3 & 2 & 300 + 300 & - & - & - & -\\
      & 4 & 3 & 300 + 300 & - & - & - & -\\
      & 4 & 4 & 300 + 300 & - & - & - & -\\
      & 4 & 5 & 300 + 300 & - & - & - & -\\
      & 5 & 6 & 300 + 300 & - & - & - & -\\
      & 5 & 7 & 300 + 300 & - & - & - & -\\
      & 7 & 8 & 300 + 300 & - & - & - & -\\
      & 7 & 9 & 300 + 300 & - & - & - & -\\
      & 8 & 10 & 300 + 300 & - & - & - & -\\
      & 8 & 11 & 300 + 300 & - & - & - & -\\
      \midrule
      \multirow{3}{*}{\textbf{Validation set}}
      & 1 & 12 & 300 + 300 & 300 + 300 & - & - & -\\
      & 1 & 13 & 300 + 300 & 300 + 300 & - & - & -\\
      & 1 & 14 & 300 + 300 & 300 + 300 & - & - & -\\
      \midrule
      \multirow{3}{*}{\textbf{Test set}}
      & 9 & 15 & 300 + 300 & 300 + 300 & 300 + 300 & 300 + 300 & 300 + 300\\
      & 9 & 16 & 300 + 300 & 300 + 300 & 300 + 300 & 300 + 300 & 300 + 300\\
      & 9 & 17 & 300 + 300 & 300 + 300 & 300 + 300 & 300 + 300 & 300 + 300\\
      \bottomrule
    \end{tabular}
    }
\label{tab:dataset}
\end{table*}

\subsubsection{Choice of ``Corruption'' Types}
The ``corruption'' types are selected against three criteria. First, clinical frequency: smoke from energized instruments, bleeding, and inadequate illumination are among the most common causes of visual degradation in endoscopic and robot-assisted procedures. Second, and most restrictively, physical reproducibility under kinematic replay: because our paired data depends on replaying an identical trajectory, an admissible ``corruption'' must be introducible without disturbing the robot, the calibrated camera pose, or the timing of the replay. This criterion excludes several otherwise desirable candidates -- for example, lens contamination alters the optics between replays and so breaks the pixel-level pairing that gives the benchmark its value, and motion blur would require changing trajectory speed, which changes the tool poses being compared. Third, mechanistic coverage: the selected types span three distinct ways in which an image can degrade -- appearance and background shift (background change), additive occlusion and scatter (bleeding, smoke), and signal loss (low brightness) -- rather than sampling several variations of one mechanism. Background change additionally serves as a control condition: it alters the scene without occluding or obscuring the tools, which isolates domain shift from visibility loss.

\subsubsection{Data Collection}
The data was acquired in the Robotorium of the Laboratory for Computational Sensing and Robotics (LCSR), Johns Hopkins University, by surgical robotics experts familiar with operating the da Vinci robot via dVRK~\cite{kazanzides2014open}. We use dVRK~\cite{kazanzides2014open} as our robot platform with the endoscopic camera manufactured by SCHÖLLY and the image process unit manufactured by Ikegami as the perception part. We refer to the data collection pipeline in CaRTS~\cite{ding2022carts,ding2023rethinking} to generate paired ``corrupted'' video sequences. We first adjust the camera/robot configuration and perform calibrations. For each configuration, we collect different trajectories by human operation. For each trajectory, we replay the recorded kinematics to reproduce the trajectory and record video sequences under different scenarios, including ``uncorrupted'' and ``corrupted'' versions with background, smoke, bleeding, and low brightness. To generate non-adversarial ``corruptions'' for the same testing samples, we record the robot trajectory, replay the kinematics via dVRK, and manually add ``corruption'' for each replay. The background ``corruption'' is generated by changing the type of background tissue. The smoke ``corruption'' is mimicked by adding artificial fog via a fog machine. The bleeding ``corruption'' is mimicked by fake blood. The low brightness ``corruption'' is generated by turning down the camera light. The data collection pipeline is shown in the following steps:

\begin{itemize}
\item \textbf{Step 1: Camera calibration and hand-eye calibration -} We perform a camera calibration to get the intrinsics for the left and right cameras of the stereo and a hand-eye calibration to get the transformation from both cameras to the base frame of both PSM1 and PSM2. 
\item \textbf{Step 2: Trajectory generation -} We use the teleoperation feature of the dVRK to manipulate the PSMs to generate trajectories in free space and record the kinematics of the trajectory. 
\item \textbf{Step 3: Trajectory replay and recording -} We replay the same trajectories and record the videos at 10 fps with the same robot configuration under different scenarios: (1) pure dark background samples for ground truth generation, (2) ``uncorrupted'' samples. (3) non-adversarial ``corrupted'' samples.
\end{itemize}

\subsubsection{Data Annotation}
The annotations are generated via a semi-automatic pipeline based on exclusively collected sequences with purely black backgrounds, where the only salient areas in the images are the PSMs. We first automatically generate a segmentation mask for the PSMs via the segment anything model (SAM)~\cite{KirillovMRMRGXW23SAM}, where the prompts are generated via a traditional background extraction algorithm~\cite{kaewtrakulpong2002improved}. We select experts who are familiar with the da Vinci robot and segmentation task to be the annotators to verify and correct the automatically generated annotation.  Two annotators performed this verification and correction stage. Both are graduate researchers with 3 years of surgical data science research experience, selected for familiarity with both the da Vinci system and the tool segmentation task. Each examined the same sequences independently, and the union of the two annotators' selections was taken as the set of failure cases, so that a frame flagged by either annotator was corrected; this rule is deliberately conservative, maximizing the number of frames receiving human attention rather than requiring agreement before correction.  The annotation pipeline can be expressed in the following steps:
\begin{itemize}
  \item \textbf{Step 1: Prompt generation -} We use a traditional background extraction algorithm to generate coarse masks for the foreground PSMs. The coarse masks are converted to the prompt points and bounding boxes for SAM.
  \item \textbf{Step 2: Automatic generation -} We input the prompt with the image to generate a fine mask for the robot tool. The first two steps are fully automatic.
  \item \textbf{Step 3: Failure case selection -} Human annotators are involved in examining the mask generated from SAM and selecting the failure cases. We had two annotators examining the same sequences and using the union of their selections as the set of failure cases.
  \item \textbf{Step 4: Manual correction -} The human annotators manually refine the selected failure cases from step 3 via SAM with human prompt input until satisfactory results are achieved. If the annotator had three attempts but did not get satisfactory results, they would draw the contour to annotate this sample manually.
\end{itemize}

\subsection{Evaluation}
We introduce how we quantitatively evaluate the performance of an algorithm and how we decide the ranking of multiple algorithms that participate in the challenge.
\noindent\textbf{Metrics:}
We use the DICE similarity coefficient (DSC) and normalized surface distance (NSD) averaged from different tolerances for the robot tool for multiple ``corruptions'' - low brightness, smoke, and blood. 
The DSC is a widely used metric in the field of medical image analysis. For a 2D binary class image segmentation task, DSC is defined as $$DSC=\frac{2TP}{2TP+FP+FN}$$ where $TP$ is the number of true positive pixels for the class, $FP$ is the number of false positives, and $FN$ is the number of false negatives. NSD measures the overlap of two boundaries between the predicted and ground truth segmentation masks. A boundary pixel is counted as overlapping when the closest distance to the other boundary is less than or equal to the specified tolerance. NSD for a 2D binary class image segmentation task is defined as: $$NSD=\frac{\left|S_A \cap \mathcal{B}_B^{(\tau)}\right|+\left|S_B \cap \mathcal{B}_A^{(\tau)}\right|}{\left|S_A\right|+\left|S_B\right|}$$ where $A$ and $B$ are the masks, $S_A$ and $S_B$ are boundaries, and $\mathcal{B}_A^{(\mathrm{\tau})}$ and $\mathcal{B}_B^{(\mathrm{\tau})}$ are the boundary of $A$ and $B$ with an extended border of width tolerance $\tau$. We calculate the average DSC and NSD over images. DSC is the standard metric for segmentation, while NSD is complementary to DSC. DSC reveals performance more in the chunk area, while NSD focuses on the boundary. We evaluate the performance only on the unseen domain to encourage participants to focus on the algorithm's robustness. In our experiment and challenge, we set $\tau = 5$ with the test resolution at $480 \times 270$. We report metrics to four decimal places for better differentiation of the participated team.

\subsection{Challenge Logistics}

SegSTRONG-C challenge launched as a one-time sub-challenge under the EndoVis2024 challenge\footnote{https://opencas.dkfz.de/endovis/challenges/2024/}. The challenge launched on May 1st, 2024, with all training and validation datasets released. 
All submissions were collected before September 6th, 2024, and evaluated using Docker containers. We announced one winner and two runner-ups on MICCAI2024 (October 10th, 2024). The final results were released on October 12th, 2024, and test data were released on October 18th, 2024.

\subsubsection{Challenge Objective}
Our goal is to encourage algorithms to exhibit robustness to unforeseen yet plausible complications that may arise during surgery, such as smoke, over-bleeding, and low brightness for the robot tool segmentation task.  The ``corruption'' types used for testing -- bleeding, smoke, and low brightness -- were disclosed in the challenge description at launch. What was withheld was the test data: the corresponding sequences remained unreleased until after the final results were announced, and the released training set contained no ``corrupted'' frames at all. Participants therefore knew which ``corruptions'' they would face.

\subsubsection{Registration and Submissions}
We provide open registration for all participants of interest. The registration application was submitted through Google Forms along with a signed agreement to the EndoVis participation rules. We review the participation applications with signed agreements and verify and approve the participation applications from valid research institutions from both industry and academia.
In total, we received 40 registrations from all over the world. All related countries are listed below: Canada (12.5\%), USA (7.5\%), Republic of Korea (10\%), China (32.5\%), Switzerland(2.5\%), Japan(2.5\%), Portugal(2.5\%), India(10\%), Germany(7.5\%), UK(2.5\%), Iran(2.5\%), Denmark(2.5\%), Singapore(2.5\%), Morocco(2.5\%). Within all registrations, we received 10 valid submissions. They are from Canada (20\%), USA (20\%), Republic of Korea (20\%), China (10\%), Switzerland(10\%), Japan(10\%), Portugal(10\%).

\subsection{Baseline Models}\label{sec:baseline}
In this section, we introduce the selected segmentation networks and data augmentation methods for baseline analysis. 

\subsubsection{DeepLabV3+~\cite{Chen2018deeplabv3plus}}
The most recent version of the DeepLab family is DeepLabV3+. The architecture is changed to an encoder-decoder structure. The encoder part makes use of atrous separable convolution, which separates the convolution process into a Depthwise Convolution and a Pointwise Convolution. Depthwise Convolution applies different kernels to each channel, and Pointwise Convolution combines the outputs across channels. The decoder part combines the low-level features and the upsampled output from the encoder to recover the spatial information by upscaling. 

\subsubsection{UNet++~\cite{zhou2019unetplusplus}}
UNet++ is a nested, deeply supervised encoder-decoder architecture designed to enhance the segmentation performance of the standard U-Net. Its core innovation lies in the introduction of redesigned skip pathways that feature dense, nested convolutional blocks connecting the encoder and decoder sub-networks. By capturing multi-scale features at varying depths, these dense connections effectively bridge the semantic gap between the feature maps of the contracting and expanding paths, thereby enabling more precise segmentation of fine anatomical details in medical imaging.

\subsubsection{nnUNet~\cite{isensee2021nnu}}
nnU-Net ("no-new-Net") is a self-configuring deep learning framework that achieves state-of-the-art performance by automating the adaptation of standard 2D and 3D U-Net architectures to specific medical datasets. Rather than proposing novel layers, it systematically analyzes the "fingerprint" of the training data—such as voxel spacing and intensity distributions—to dynamically determine the optimal preprocessing, network topology, and training hyperparameters. This systematic approach effectively eliminates the need for manual tuning and heuristic architectural design, establishing nnU-Net as the robust benchmark in medical image segmentation challenges.

\subsubsection{SegFormer~\cite{xie2021segformer}}
SegFormer is a semantic segmentation framework that combines a hierarchical transformer encoder with a multi-layer perceptron (MLP) decoder. The hierarchically structured transformer encoder avoids the interpolation of positional codes and outputs multi-scaled features. The lightweight MLP decoder combines both local and global attention to aggregate the multi-layer information. 

\subsubsection{SETR~\cite{Zheng2021SETR}}
SETR provides an alternative perspective to the encoder-decoder-based FCN architecture. It treats semantic segmentation as a sequence-to-sequence prediction task in which the author splits the image into patches, linearly projects each patch, adds positional embeddings, and feeds the sequence into the transformer encoder. Three different decoders are designed to perform pixel-level segmentation, and we implement all three: (1) Naive upsampling which projects the transformer feature to the dimension of class number and directly upsample to the desired image resolution, (2) Progressive upsampling which alternates convolution and upsampling operations instead of one-step upscaling, and (3) Multi-Level feature aggregation which takes feature representations from $M$ transformer layers uniformly, reshapes and aggregates the feature map top-down, fuses features from all channels and upsamples to the full resolution. We apply naive upsampling in our baseline analysis.

\subsubsection{Mask2Former~\cite{cheng2021mask2former}}
Mask2Former is a universal image segmentation architecture capable of unifiedly addressing semantic, instance, and panoptic segmentation tasks. Its key innovation is the masked attention mechanism, which extracts localized features by constraining cross-attention within the Transformer decoder to predicted mask regions rather than the full image context. By coupling this efficient attention mechanism with a multi-scale pixel decoder, Mask2Former significantly improves convergence speed and segmentation accuracy, effectively generalizing across diverse segmentation benchmarks without requiring task-specific architectural modifications.

\subsubsection{SAM2~\cite{ravi2024sam}}
SAM2 extends the Segment Anything Model (SAM) for video segmentation by incorporating temporal information. It retains SAM’s core components—image encoder, prompt encoder, and mask decoder—while introducing a memory encoder and memory bank to capture frame-to-frame consistency. The memory mechanism allows SAM2 to leverage past predictions, enhancing temporal coherence in segmentation. We adopt SAM2 as a baseline for its strong performance in point prompt-based video segmentation and its ability to model temporal context effectively. We use the ground truth mask to sample 5 positive points and 5 negative points for the first frame of each sequence.

\subsubsection{SAM3~\cite{carion2025sam}}
SAM 3 extends the architecture of SAM 2 by introducing Promptable Concept Segmentation (PCS), moving beyond single-object tracking to open-vocabulary concept identification. It replaces the separate image and prompt encoders with a unified Perception Encoder and incorporates a Presence Head to decouple recognition from localization. This unified framework allows SAM 3 to detect, segment, and track all instances of a semantic category (e.g., via text descriptions or visual exemplars) throughout a video sequence with maintained temporal consistency. We adopt SAM 3 as a baseline to evaluate the performance of semantic-driven segmentation, utilizing either point prompts sampled from the ground truth mask or a text prompt describing the instrument (Section~\ref{sec:results_baseline}) to initialize the tracking.

\subsubsection{SurgSAM2~\cite{liu2024surgical}}
SurgSAM-2 adapts the SAM 2 framework specifically for the resource-constrained environment of real-time surgery. While retaining the core SAM 2 architecture, it introduces an Efficient Frame Pruning (EFP) mechanism to optimize the memory bank. Instead of naively storing all past frames, SurgSAM-2 dynamically evaluates frame importance via cosine similarity, selectively retaining only the most informative features while discarding redundant temporal data. This approach significantly reduces computational overhead—achieving a threefold increase in inference speed—while maintaining high-fidelity segmentation performance on surgical benchmarks.

\subsubsection{YOLOv11~\cite{redmon2016you,khanam2024yolov11}}
YOLO11 represents a significant evolution in the You Only Look Once (YOLO) family, engineered by Ultralytics to redefine the efficiency-accuracy trade-off in real-time detection and segmentation. It introduces a refined backbone architecture incorporating C3k2 blocks for enhanced feature extraction and C2PSA (Cross-Stage Partial with Spatial Attention) blocks to optimize spatial awareness with minimal computational cost. These architectural advancements allow YOLO11 to achieve superior parameter efficiency compared to its predecessors (e.g., YOLOv8), making it an ideal candidate for high-speed, resource-constrained surgical applications where low latency is critical.

\subsubsection{AutoAugment~\cite{cubuk2019autoaugment}}
AutoAugment searches for the best sequence of transformations that improves the performance of a model on a given dataset. We modify the Pytorch implementation of the AutoAugmentPolicy class and choose the policy learned on ImageNet. We apply AutoAugment during the training of UNet for this baseline.

\subsubsection{Customized Synthetic ``Corruption'' Augmentation}
To address domain shifts caused by environmental ``corruptions,'' we designed a \textit{Customized Synthetic ``Corruption'' Augmentation} pipeline inspired by the most effective strategies identified in the SegSTRONG-C benchmark. First, to mitigate the impact of bleeding, we employed a geometric proxy method that generates random red ellipses and general occlusion masks to simulate visual obstruction without relying on computationally expensive fluid dynamics. Second, for smoke simulation, we implemented a procedural volumetric effect utilizing the Diamond-Square algorithm to generate continuous, cloud-like noise maps, which are alpha-blended with the training images at stochastic densities. Finally, we integrated pixel-level jittering to systematically reduce brightness and contrast, explicitly conditioning the model to generalize across underexposed surgical scenes.

\subsection{Challenge Submissions}
\label{sec:submissions}

\begin{figure}[ht]
\begin{center}
    \includegraphics[width=\linewidth]{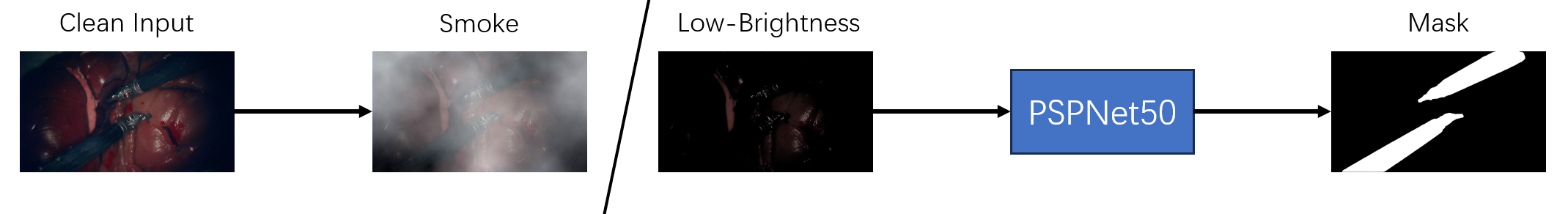}
\end{center}
   \caption{Overview of our proposed ``corruption'' simulation + segmentation framework. (Team VSI).
   }
\label{fig:vsi_fig1}
\end{figure}

\subsubsection{VSI (Yicheng Leng, Seok Bong Yoo, Eung-Joo Lee)}
Team VSI's framework is illustrated in Figure~\ref{fig:vsi_fig1}. They use PSPNet50 ~\cite{zhao2017pyramid} for the surgical tool segmentation. To improve the model's robustness against different types of ``corruption,'' they apply image augmentation techniques from ImageNet-C \cite{hendrycks2019benchmarking} to simulate the smoke and low-brightness cases. They ignore the augmentation of the bleeding case as they think the color and shape of red artifacts added onto the image are difficult to determine, and the bleeding area brings minimal pixel-wise impact on both the surgical tool and tissue background.

\subsubsection{UBCRCL (Zijian Wu, Yuxin Chen, Septimiu E. Salcudean)}
RGB-D semantic segmentation methods have been demonstrated to improve the segmentation performance by fusing depth information~\cite{jamal2024rethinking}. Team UBCRCL tries to explore the performance of an RGB-D method on surgical datasets, especially when confronting ``corrupted'' images.
They adopt a state-of-the-art method, DFormer~\cite{yin2023dformer}, a pretraining framework to encode RGB-D representations. They apply an off-the-shelf DepthAnything~\cite{yang2024depth} model to generate pseudo depth for each frame, as the depths from stereo-matching~\cite{lipson2021raft} are not satisfying without rectified images. To enhance the model's robustness, they perform the 3D Common ``Corruption'' (3DCC)~\cite{kar20223d} for data augmentation. Specifically, they tune the following augmentation to different degrees during the training: reducing the image brightness, the color and density of fog, and random occlusions.

\begin{figure}[ht]
\includegraphics[width=\linewidth]{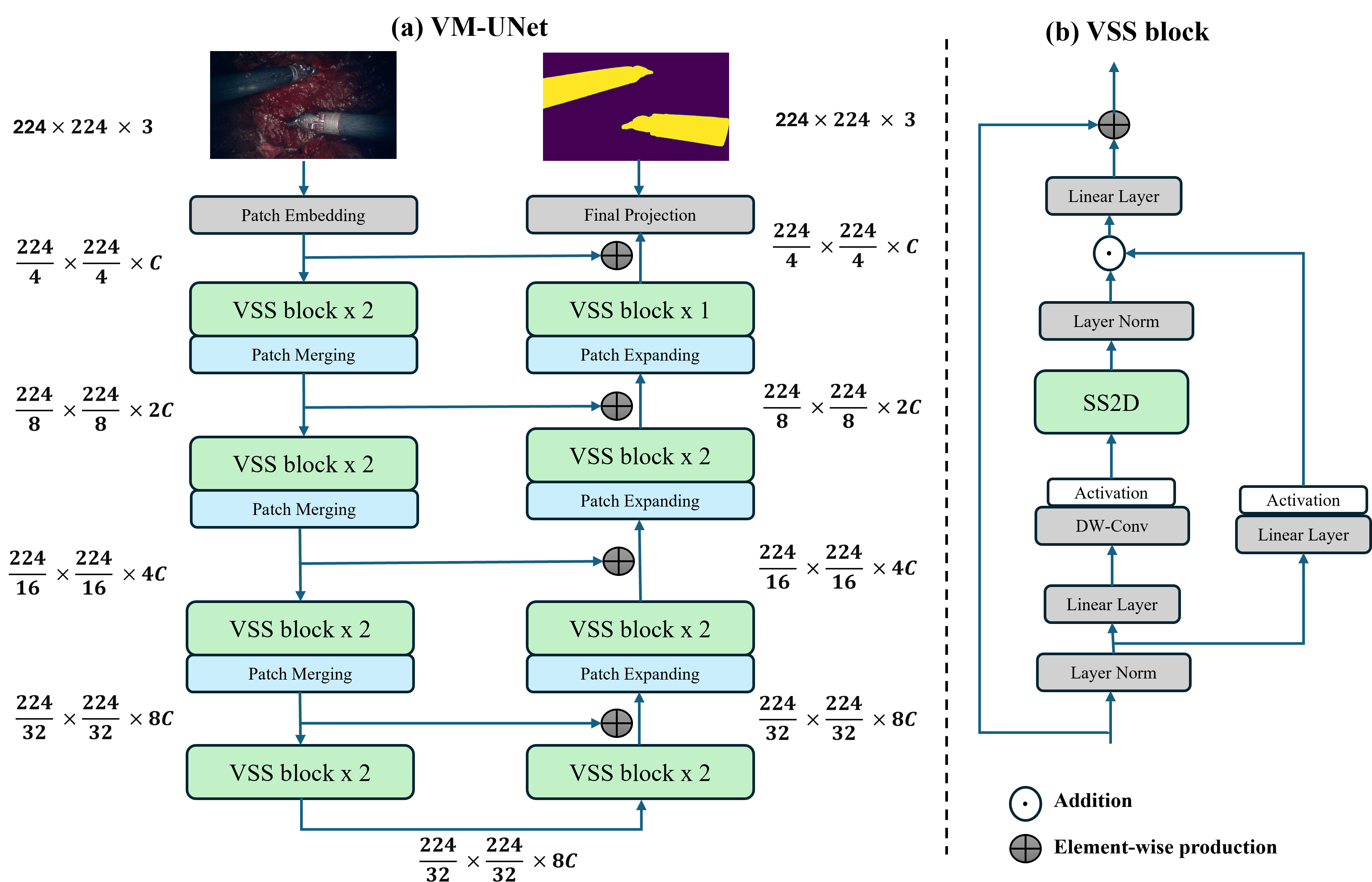}
\caption{(a). VM-UNet architecture. (b) Details of components in VSS block (Team SAM\_KHU).} \label{fig:SAM_KHU_fig1}
\end{figure}

\begin{figure}[ht]
    \centering
    \includegraphics[width=\linewidth]{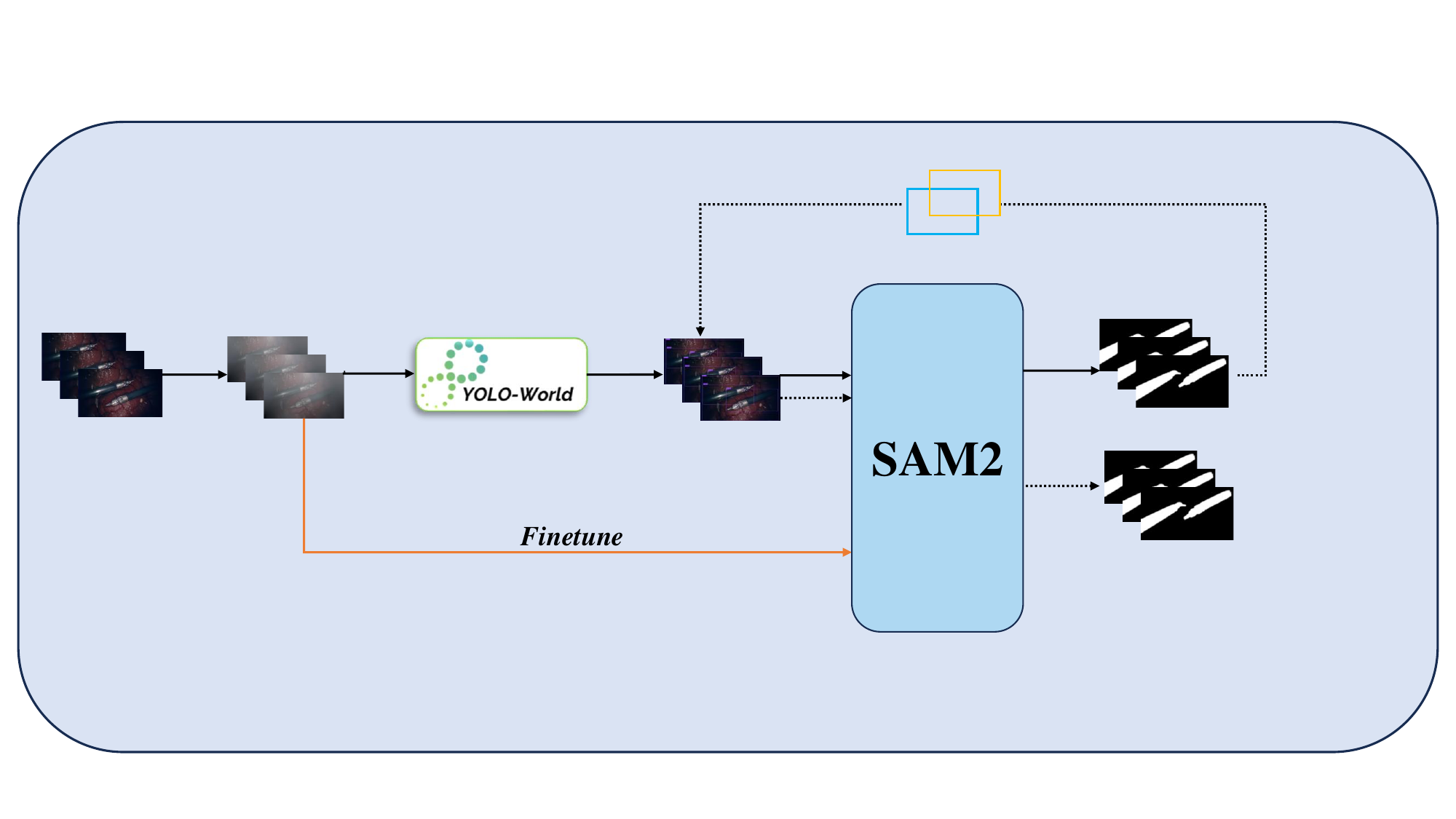}
    \caption{General workflow of our methodology. (Team Medical\_Mechatronics).}
    \label{fig:Medical_Mechatronics_flow}
\end{figure}

\begin{figure*}[t]%
\centering
\includegraphics[width=1\textwidth]{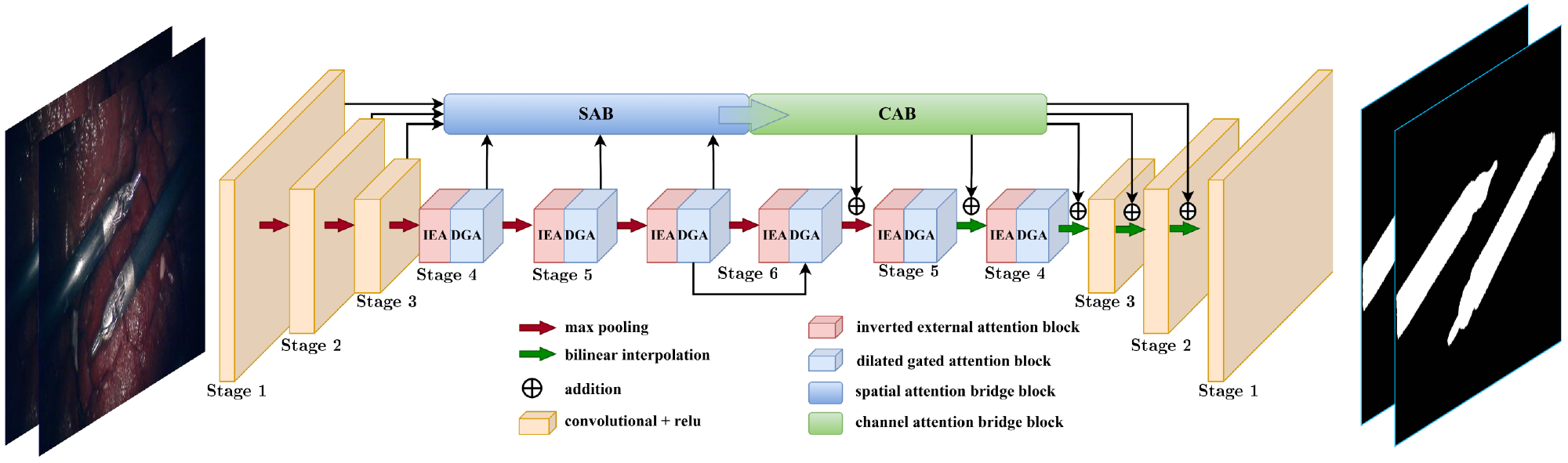}
\caption{The overall architecture of the proposed surgical tool segmentation model. (Team CCG\_DGIST).}\label{fig:CGG_fig1}
\end{figure*}

\begin{figure}[ht]
    \centering
    \begin{subfigure}[h]{\linewidth}
        \centering
        \includegraphics[width=\linewidth]{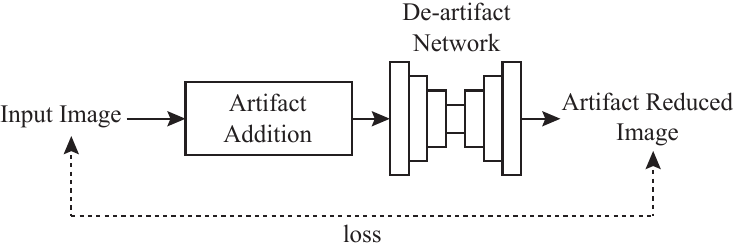}
    \end{subfigure}
    \begin{subfigure}[h]{\linewidth}
        \centering
        \includegraphics[width=\linewidth]{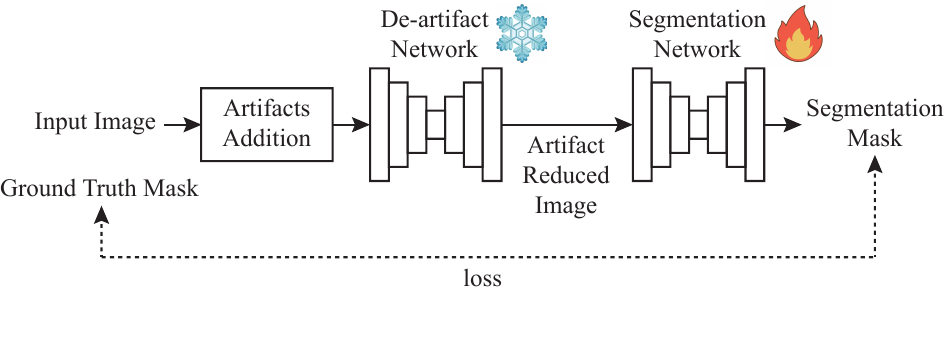}
    \end{subfigure}
    \caption{Training the de-artifact network and the segmentation network. (Team SK).}
    \label{fig:sk_network}
\end{figure}

\begin{figure}
    \centering
  \includegraphics[width=\linewidth]{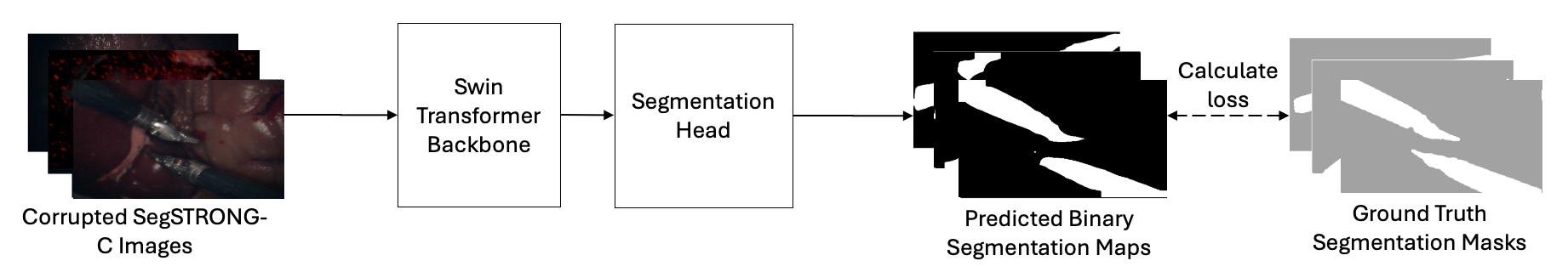}
  \caption{Overview of the model and training. (Team Tailor).}
  \label{fig:tailor_framework}
\end{figure}

\subsubsection{SAM\_KHU (Samra Irshad, Shadi Albarqouni, Seong Tae Kim)}
State space models (SSMs) \cite{gu2023mamba,kalman1960new} have recently gained attention for their ability to model long-distance dependencies with linear complexity, improving sequence modeling and computational efficiency, and are now being integrated with architectures like CNNs for tasks such as medical image segmentation \cite{ma2024u, xing2024segmamba}. Team SAM\_KHU applies Vision Mamba-UNet (VM-UNet), a recently proposed SSM-based model for this challenge. The architecture is shown in Figure~\ref{fig:SAM_KHU_fig1}. The encoder includes VSS blocks from VMamba \cite{liu2024vmamba} for feature extraction and patch merging operations for downsampling. The decoder employs VSS blocks and patch-expanding operations to recover the original size of the segmentation output. The skip connection module uses a simple additive operation.  To make the model robust to procedural artifacts, they use StarGanv2 \cite{choi2020stargan}, an image-to-image translation model that learns the mapping between source and target domains to synthesize these patterns.

\subsubsection{Medical\_Mechatronics (Yueyi Sun, An Wang, Long Bai, Hongliang Ren)}
For Team Medical\_Mechatronics, their primary objective is to introduce comprehensive perturbations onto the pristine images, effectively synthesizing a new, realistic dataset with characteristics mimicking smoke, bleeding, and low-light conditions. By training on this augmented dataset, they aim to develop a more robust model that can generalize well to unseen test data. They have created different sets of datasets for each ``corruption'' and trained type-specific models. For smoke, they leverage the widely used fog synthesis approach from the robustness benchmark~\cite{hendrycks2019benchmarking} to generate ``corruption'' at different severity levels. For blood, they employ the implementation from the semi-synthetic approach~\cite{garcia2021image}, utilizing the \textit{BloodDroplet} class to introduce blood droplets onto images. For low brightness, they utilize Photoshop's batch processing capabilities to enhance images and achieve a similar effect. They leverage the great generalization ability of pre-trained vision foundation models in their design. Specifically, they employ Yolo-World~\cite{cheng2024yolo} to generate bounding box prompts for Segment Anything Model 2 (SAM 2)~\cite{ravi2024sam}, which will be fine-tuned with their augmented datasets. The overall pipeline of their method is demonstrated in Figure~\ref{fig:Medical_Mechatronics_flow}.

\subsubsection{CCG\_DGIST (Ihsan Ullah, Ho-Gun Ha, Attaullah Khan, Hyunki Lee)}
Team CCG\_DGIST utilizes a multi-attention and lightweight model that achieves competitive performance in surgical tool segmentation. The overall framework is shown in Figure~\ref{fig:CGG_fig1}. This framework incorporates several attention modules to enhance performance: (1) The Dilated Gated Attention Block (DGA), which enables the model to focus more precisely on the target regions by capturing both global and local information. (2) The Inverted External Attention Block (IEA)\cite{guo2022beyond, sandler2018mobilenetv2}, an efficient external attention mechanism that strengthens the information associations between samples and extracts overarching characteristics from the entire dataset. (3) To capture multistage and multiscale information, they introduce attention bridge modules for both channel and spatial dimensions, termed the Channel Attention Bridge Block (CAB) and Spatial Attention Bridge Block (SAB)\cite{woo2018cbam}, which generate their respective attention maps. Additionally, depthwise separable convolutions~\cite{howard2017mobilenets} are employed to reduce the number of parameters and computational complexity, making the model more efficient.

\subsubsection{SK (Satoshi Kondo, Satoshi Kasai, Kousuke Hirasawa)}
Team SK applies a two-step regime for the challenge. They use two networks: a de-artifact network and a segmentation network. The top subfigure of Figure~\ref{fig:sk_network} shows the de-artifact network. They add one of the three artifacts: brightness, smoke, and blood to the input image for training. The brightness artifacts are realized by reducing the brightness and contrast in a predefined range. The smoke artifacts are realized by alpha-blending the input and artificial smoke image. The blood artifacts are realized by adding red ellipses to the input image. The de-artifact network is an encoder-decoder architecture like UNet~\cite{RonnebergerFB15unet}. The encoder part is replaced by ConvNeXt-v2-base~\cite{woo2023convnext}. The loss function is the L1 loss between the input image and the artifact-reduced image. The bottom subfigure of Figure~\ref{fig:sk_network} shows the segmentation network. The artifact addition block has the same functionality as the one when the de-artifact network is trained. For robustness, artifact addition is applied to randomly selected 75 \% of the training datasets. The de-artifact network trained in the first step is frozen, and only the segmentation network is trained. The architecture of the segmentation network is the same as the de-artifact network. The loss function is a weighted summation of Dice loss, cross-entropy loss, and Hausdorff distance loss.

\subsubsection{Tailor (Sita Tailor, Ricardo Sánchez-Matilla, Imanol Luengo)}
Team Tailor applies both surgical-specific and general augmentations to better simulate real surgical conditions. The surgical-specific augmentations included motion blur, low lighting, blood, and smoke, mimicking non-adversarial ``corruptions'' typical in operating environments. Additionally, general augmentations like affine transformations, flips, and random resized cropping are applied to introduce further variability. All augmentations were implemented using functions from the Albumentations library \cite{buslaev2020albumentations}. They focus on transformer-based models for robust surgical tool segmentation, selecting the SwinV2 Transformer (SwinV2 small) \cite{liu2022swin} for its superior performance with segmentation tasks. They modify the model by replacing the classification head with a segmentation head, incorporating a 1x1 convolutional layer followed by upsampling to generate segmentation maps. An overview of the framework is displayed in Figure~\ref{fig:tailor_framework}.

\subsubsection{FightTumor (Tianhao Fu, Jun Ma, Bo Wang)}
Team FightTumor's approach uses pre-trained UNet~\cite{RonnebergerFB15unet} with AutoAugment~\cite{cubuk2019autoaugment}. They customize augmentations in Albumentations \cite{buslaev2020albumentations}. For smoke, the randomized smoke image is merged on top of the original image with an opacity of 1. They also generate a circle flare in which the border is blurred.

\subsubsection{ICVS-2AI (Marcos Fernández-Rodríguez, Estevão Lima, João L. Vilaça)}
For Team ICVS-2AI, a CycleGAN is trained with default parameters~\cite{zhu2017CycleGAN} and unpaired images, chosen from a private laparoscopic dataset, with 1496 clear and 220 containing smoke artifacts. To learn the “clear to smoke” style, they use the trained model with the SegSTRONG-C dataset, generate smoke versions of the original images, and use the original ground truth for the new images. Finally, the nnUNet network is trained using standard parameters described in their paper~\cite{isensee2021nnu}, with the increased dataset, using a combined Dice and top-k loss.

\subsubsection{AIMI (Negin Ghamsarian, Klaus Schoeffmann, Raphael Sznitman)}
Team AIMI explore three architectures ReCalNet~\cite{ghamsarian2021recal}, AdaptNet~\cite{ghamsarian2021lensid}, and DeepPyramid+~\cite{ghamsarian2024deeppyramid+,ghamsarian2022deeppyramid}. After thorough experiments, they finally applied DeepPyramid+~\cite{ghamsarian2024deeppyramid+}, which features two main modules: Pyramid View Fusion (PVF) and Deformable Pyramid Reception (DPR) as their architecture. The PVF module enhances pixel-level information representation by simulating the human visual system, offering a global-to-local perspective to capture relevant details around each pixel. The DPR module employs dilated deformable convolutions to facilitate shape- and scale-adaptive feature extraction, thereby improving segmentation accuracy for diverse and deformable classes within medical images. They apply common data augmentations like random cropping, color jittering, grayscale transformation, and Gaussian blur during training for better robustness.

\section{Results and Discussion}
\subsection{Baseline Analysis}\label{sec:results_baseline}
To establish reference performance and reveal fundamental challenges in robust segmentation under ``corruption,'' we conducted a baseline analysis using carefully selected representative models. We included UNet++\cite{zhou2019unetplusplus}, nnUNet~\cite{isensee2021nnu} and DeepLabv3plus\cite{Chen2018deeplabv3plus} as popular choice examples of CNNs and SETR\cite{Zheng2021SETR}, Segformer\cite{xie2021segformer}, and Mask2Former~\cite{cheng2021mask2former} as representatives of transformer-based architectures. All models were trained in a fully supervised manner on ``uncorrupted'' training sequences in line with the challenge requirements. 
We apply AutoAugment~\cite{cubuk2019autoaugment} during the training of all baselines.

In addition, we randomly sample point prompts from ground truth for the SAM2~\cite{ravi2024sam}, SAM3~\cite{carion2025sam}, and SurgSAM2~\cite{liu2024surgical}. We also applied the text prompt ``Surgical tool, usually silver needle drivers, grasper, hook, or scissors'' for SAM3.

The results, detailed in Table~\ref{tab:baseline}, present the mean performance and standard deviation for each baseline. To improve statistical reliability, metrics are averaged over five independent runs, utilizing distinct random seeds for the training of end-to-end DNNs and for point-prompt sampling in SAM-based models across the various ``corruption'' types~\cite{christodoulou2025false,christodoulou2024confidence}. On the ``uncorrupted'' test set, all representative end-to-end DNNs demonstrate robust capability, achieving Dice Similarity Coefficients (DSC) exceeding 0.9 and Normalized Surface Distances (NSD) above 0.8. However, their performance degrades when evaluated on sequences subject to environmental ``corruptions,'' including background change, bleeding, smoke, and low brightness. This empirically validates that even naturally occurring, non-adversarial perturbations cause substantial deterioration in the efficacy of end-to-end trained networks.
Similarly, SAM-based models exhibit performance declines in the presence of bleeding and smoke and fail substantially in low-brightness scenarios. Notably, SurgSAM2—despite being fine-tuned on surgical data—underperforms compared to the baseline SAM2. We hypothesize that this discrepancy stems from a significant domain shift between the external fine-tuning datasets and our target distribution; consequently, the fine-tuning process appears to compromise the foundation model's zero-shot generalization without conferring benefits for our specific domain. 
\textbf{The consistent performance drops observed across all baseline architectures under ``corrupted'' conditions underscore the critical robustness limitations inherent in current model development practices.} 

Table~\ref{tab:baseline} also reports how each baseline was initialized, and includes UNet++ trained from scratch. Removing pretraining costs UNet++ $0.0145$ DSC on the ``uncorrupted'' domain and $0.0762$ on the ``corrupted'' average, moving it from the strongest entry in the table to fourth of the five from-scratch configurations. Its lead is therefore substantially an initialization effect rather than an architectural one. Initialization is not the only route to it, however: nnUNet uses no pretrained weights, and its default configuration is the second strongest end-to-end baseline on the ``corrupted'' average, above the pretrained DeepLabv3+ and Mask2Former. Since nnUNet also self-configures its preprocessing, augmentation and schedule, this indicates that a training recipe tuned to the dataset can substitute for part of what pretraining supplies.

\begin{table*}[ht]
  \centering
  \caption{Experiment results of baseline models across different domains. Values are mean $\pm$ standard deviation over five independent runs (distinct training seeds for the end-to-end networks; distinct prompt-sampling seeds for the SAM-family baselines). The SAM3 text-prompt rows carry no standard deviation because that configuration is deterministic given a fixed prompt and was run once. Init.: whether the model was initialized from pretrained weights or trained from scratch. The nnUNet rows separate the augmentation pipeline used in the challenge baseline (AutoAugment) from nnUNet's own default augmentation, which is the configuration a reader would obtain by running nnUNet unmodified.}
  \resizebox{\textwidth}{!}{
    \begin{tabular}{c|c|c|c|c|c|c}
      \toprule
      \multirow{ 2}{*}{\textbf{DSC Score}}  & \multirow{ 2}{*}{\textbf{Init.}} & \multirow{ 2}{*}{\textbf{``uncorrupted''}} & \textbf{Background} & \multirow{ 2}{*}{\textbf{Bleeding}} & \multirow{ 2}{*}{\textbf{Smoke}} & \textbf{Low}  \\
      & & & \textbf{Change} & & &  \textbf{Brightness} \\
      \midrule
      nnUNet~\cite{isensee2021nnu}
      & Scratch & 0.9697 $\pm$ 0.0007 & 0.9449 $\pm$ 0.0084 & 0.7521 $\pm$ 0.0063 & 0.8836 $\pm$ 0.0155 & 0.6776 $\pm$ 0.0450 \\
      nnUNet~\cite{isensee2021nnu} Default Augmentation
      & Scratch & 0.9698 $\pm$ 0.0008 & 0.9555 $\pm$ 0.0054 & 0.7563 $\pm$ 0.0093 & 0.8798 $\pm$ 0.0050 & 0.7251 $\pm$ 0.0569 \\
      Segformer~\cite{xie2021segformer}
      & Scratch & 0.9345 $\pm$ 0.0022 & 0.9029 $\pm$ 0.0024 & 0.7103 $\pm$ 0.0101 & 0.8117 $\pm$ 0.0177 & 0.7971 $\pm$ 0.0295 \\
      SETR~\cite{Zheng2021SETR}
      & Scratch & 0.9076 $\pm$ 0.0024 & 0.8740 $\pm$ 0.0006 & 0.6610 $\pm$ 0.0078 & 0.8029 $\pm$ 0.0115 & 0.6988 $\pm$ 0.0062 \\
      DeepLabv3+~\cite{Chen2018deeplabv3plus} 
      & Pretrained & 0.9399 $\pm$ 0.0030 & 0.9192 $\pm$ 0.0048 & 0.8224 $\pm$ 0.0200 & 0.8526 $\pm$ 0.0073 & 0.7001 $\pm$ 0.0501 \\
      UNet++~\cite{zhou2019unetplusplus} 
      & Pretrained & 0.9704 $\pm$ 0.0009 & 0.9457 $\pm$ 0.0224 & 0.9306 $\pm$ 0.0371 & 0.8663 $\pm$ 0.0371 & 0.7823 $\pm$ 0.0239 \\
      UNet++~\cite{zhou2019unetplusplus}
      & Scratch & 0.9559 $\pm$ 0.0029 & 0.9299 $\pm$ 0.0034 & 0.8134 $\pm$ 0.0214 & 0.8630 $\pm$ 0.0130 & 0.6136 $\pm$ 0.0586 \\
      Mask2Former~\cite{cheng2021mask2former}
      & Pretrained & 0.9579 $\pm$ 0.0036 & 0.9226 $\pm$ 0.0088 & 0.7501 $\pm$ 0.0280 & 0.8667 $\pm$ 0.0182 & 0.7406 $\pm$ 0.0491 \\
      SAM2~\cite{ravi2024sam}  (point prompt)  
      & Pretrained & 0.8907 $\pm$ 0.0005 & 0.9190 $\pm$ 0.0002 & 0.8577 $\pm$ 0.0006 & 0.7303 $\pm$ 0.0017 & 0.4631 $\pm$ 0.0011\\
      SurgSAM2~\cite{liu2024surgical} (point prompt)  
      & Pretrained & 0.8115 $\pm$ 0.0008 & 0.8336 $\pm$ 0.0006 & 0.6591 $\pm$ 0.0008 & 0.7235 $\pm$ 0.0013 & 0.4856 $\pm$ 0.0011 \\
      SAM3~\cite{carion2025sam} (point prompt)  
      & Pretrained & 0.9001 $\pm$ 0.0011 & 0.8585 $\pm$ 0.0022 & 0.8603 $\pm$ 0.0005 & 0.8148 $\pm$ 0.0013 & 0.5763 $\pm$ 0.0023 \\
      SAM3~\cite{carion2025sam} (text prompt)
      & Pretrained & 0.6127 & 0.4450 & 0.5122 & 0.4046 & 0.0022 \\
      \midrule
       \multirow{ 2}{*}{\textbf{NSD Score}}  & \multirow{ 2}{*}{\textbf{Init.}} & \multirow{ 2}{*}{\textbf{``uncorrupted''}} & \textbf{Background} & \multirow{ 2}{*}{\textbf{Bleeding}} & \multirow{ 2}{*}{\textbf{Smoke}} & \textbf{Low}  \\
      & & & \textbf{Change} & & &  \textbf{Brightness} \\
      \midrule
      nnUNet~\cite{isensee2021nnu}
      & Scratch & 0.9794 $\pm$ 0.0049 & 0.9080 $\pm$ 0.0097 & 0.6174 $\pm$ 0.0119 & 0.8016 $\pm$ 0.0168 & 0.5619 $\pm$ 0.0264 \\
      nnUNet~\cite{isensee2021nnu} Default Augmentation
      & Scratch & 0.9782 $\pm$ 0.0035 & 0.9343 $\pm$ 0.0182 & 0.6296 $\pm$ 0.0072 & 0.8037 $\pm$ 0.0126 & 0.6053 $\pm$ 0.0432 \\
      Segformer~\cite{xie2021segformer}
      & Scratch & 0.9048 $\pm$ 0.0063 & 0.8061 $\pm$ 0.0085 & 0.5504 $\pm$ 0.0088 & 0.6987 $\pm$ 0.0109 & 0.6135 $\pm$ 0.0367 \\
      SETR~\cite{Zheng2021SETR}
      & Scratch & 0.8021 $\pm$ 0.0107 & 0.6749 $\pm$ 0.0045 & 0.4644 $\pm$ 0.0114 & 0.6154 $\pm$ 0.0182 & 0.4838 $\pm$ 0.0133 \\
      DeepLabv3+~\cite{Chen2018deeplabv3plus} 
      & Pretrained & 0.8890 $\pm$ 0.0159 & 0.8232 $\pm$ 0.0148 & 0.6582 $\pm$ 0.0244 & 0.7303 $\pm$ 0.0107 & 0.4724 $\pm$ 0.0275 \\
      UNet++~\cite{zhou2019unetplusplus} 
      & Pretrained & 0.9801 $\pm$ 0.0027 & 0.9351 $\pm$ 0.0362 & 0.8746 $\pm$ 0.0188 & 0.8357 $\pm$ 0.0374 & 0.6719 $\pm$ 0.0423 \\
      UNet++~\cite{zhou2019unetplusplus} 
      & Scratch & 0.9326 $\pm$ 0.0081 & 0.8783 $\pm$ 0.0105 & 0.6561 $\pm$ 0.0212 & 0.7732 $\pm$ 0.0197 & 0.4722 $\pm$ 0.0528 \\
      Mask2Former~\cite{cheng2021mask2former}
      & Pretrained & 0.9523 $\pm$ 0.0100 & 0.8568 $\pm$ 0.0175 & 0.6135 $\pm$ 0.0243 & 0.7740 $\pm$ 0.0159 & 0.5866 $\pm$ 0.0415 \\
      SAM2~\cite{ravi2024sam} (point prompt)  
      & Pretrained & 0.7477 $\pm$ 0.0008 & 0.8378 $\pm$ 0.0009 & 0.6830 $\pm$ 0.0014 & 0.5310 $\pm$ 0.0012 & 0.2424 $\pm$ 0.0014 \\
      SAM3~\cite{carion2025sam} (point prompt) 
      & Pretrained & 0.7876 $\pm$ 0.0017 & 0.7039 $\pm$ 0.0034 & 0.7245 $\pm$ 0.0007 & 0.6544 $\pm$ 0.0020 & 0.3854 $\pm$ 0.0012\\
      SurgSAM2~\cite{liu2024surgical} (point prompt)  
      & Pretrained & 0.6137 $\pm$ 0.0011 & 0.6415 $\pm$ 0.0006 & 0.4778 $\pm$ 0.0005 & 0.5061 $\pm$ 0.0012 & 0.3030 $\pm$ 0.0008 \\
      SAM3~\cite{carion2025sam} (text prompt)  			
      & Pretrained & 0.5910 & 0.4292 & 0.5103 & 0.3826 &  0.0024 \\
      \bottomrule
    \end{tabular}
    }
\label{tab:baseline}
\end{table*}

\begin{figure}[ht]
\centering
\includegraphics[width=\linewidth]{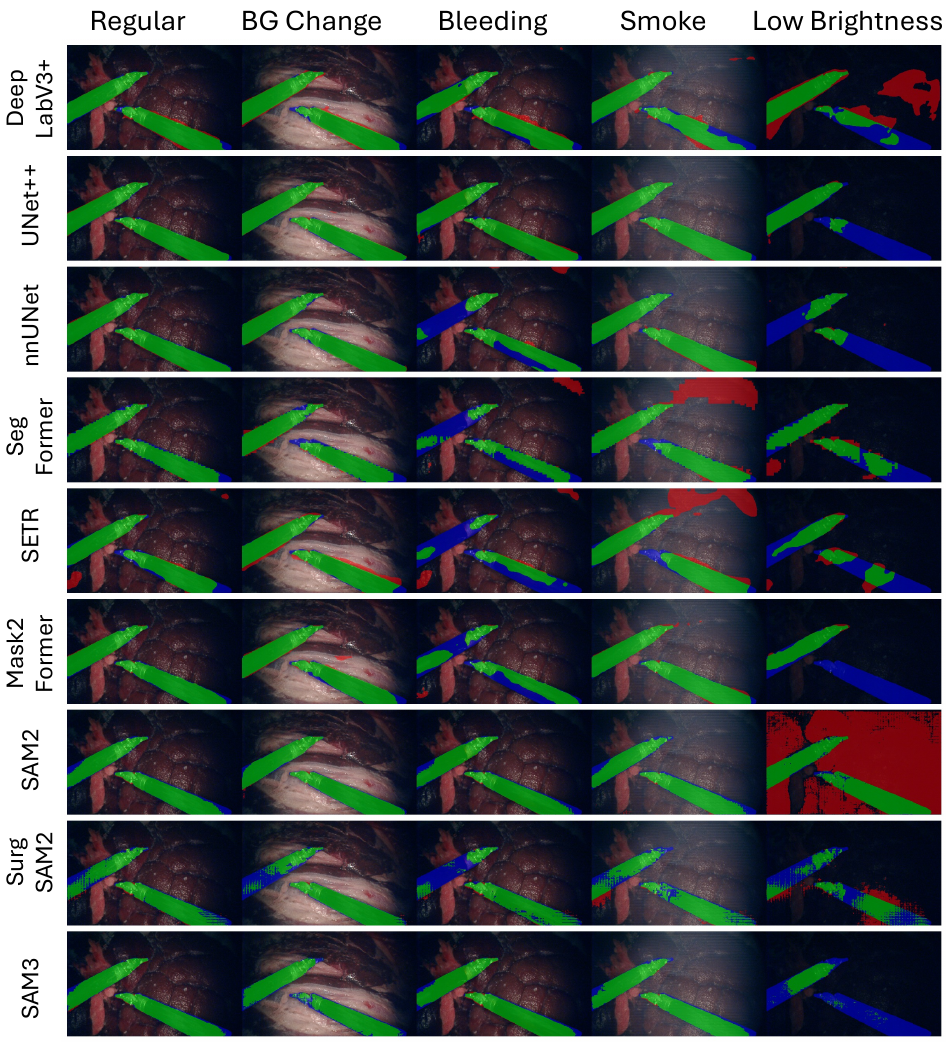}
\caption{Qualitative comparison among baseline models. The area with a green mask denotes a true positive. The red area denotes false positives. The blue area denotes false negatives.}
\label{fig:qualitative}
\end{figure}

\noindent\textbf{Failure analysis.}
We conducted a failure analysis of the baseline models to better understand the causes of performance degradation under non-adversarial ``corruptions.''
We begin with a qualitative assessment, as illustrated in the examples shown in Figure~\ref{fig:qualitative}. Each ``corruption'' type introduces distinct failure modes. In the background change domain, the appearance of the robot tool remains similar to the ``uncorrupted'' domain. Under smoke ``corruption,'' models sometimes misclassify translucent white areas introduced by smoke as surgical tools, resulting in false positives. In contrast, bleeding occludes parts of the tool, leading to false negatives. Low brightness reduces overall visibility and causes both false positive and false negative predictions.

Based on these observations -- and noting that baseline models were trained only on ``uncorrupted'' data -- we hypothesize that \textbf{performance degradation is driven by two major factors: (i) domain shift introduced by ``corruption,'' and (ii) increased task difficulty, arising from more complex background, occlusions, reduced contrast, and visual ambiguity.}
To disentangle these effects, we designed a controlled experiment specifically aimed at isolating the impact of domain shift. We selected the top-performing representatives from the CNN (UNet++) and Transformer (Mask2Former) architectures and retrained them under three distinct protocols, each incorporating varying degrees of auxiliary data from the target domains.

The experimental variants are defined as follows:
\begin{itemize}
    \item Domain-Specific Training: Models are trained exclusively on samples from the same ``corruption'' domain as the target test set.
    \item All-Domain Training: Models are trained on all ``uncorrupted'' and ``corrupted'' domains together.
    \item Incremental Domain Adaptation: Models initially trained on regular data are progressively updated with samples from either a single specific ``corruption'' domain or all available domains.
\end{itemize}

The results presented in Table~\ref{tab:ood} reveal distinct behavioral patterns across different ``corruption'' types. In the bleeding and smoke domains, models trained on either the target domain or the comprehensive dataset achieve performance comparable to the ``uncorrupted'' baseline, exhibiting only marginal deviations. This indicates that in these scenarios, the distributional shift of the training data is the primary cause of performance degradation. In contrast, within the low-brightness domain, a noticeable performance gap persists relative to the ``uncorrupted'' baseline, even when the models are explicitly trained on the target domain or on all domains together. This discrepancy is particularly pronounced in the Mask2Former architecture. These findings suggest that in low-brightness scenarios, model failure is driven by a compound effect: the domain shift introduced by the ``corruption,'' combined with increased intrinsic task difficulty (i.e., the loss of critical visual information).

\begin{figure*}[ht]
\centering
\includegraphics[width=\linewidth]{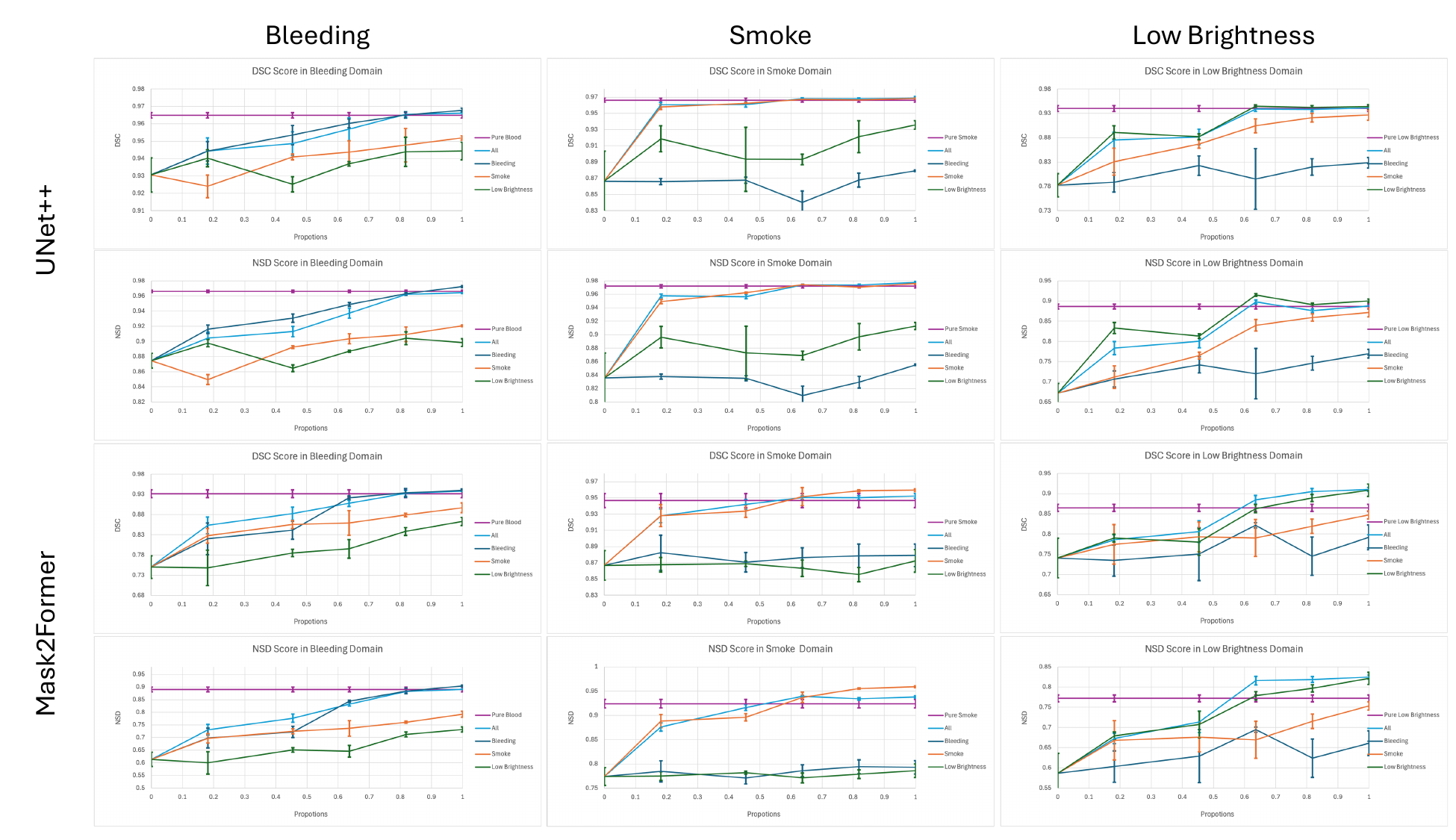}
\caption{Auxiliary data study results. Each column represents the test domain. The x-axis in each plot denotes the ratio of the auxiliary data and ``uncorrupted'' data. The y-axis denotes the DSC/NSD scores. Lines in plots refer to adding auxiliary data from different domains.}
\label{fig:aux_data}
\end{figure*}

The visualization results presented in Figure~\ref{fig:aux_data} demonstrate that incorporating training data from either the target domain or the full domain spectrum effectively neutralizes the domain gap. Crucially, \textbf{the volume of auxiliary data required to achieve this convergence varies by ``corruption'' type}. For instance, in the smoke domain, auxiliary data equivalent to only 20\% of the ``uncorrupted'' dataset size is sufficient to bridge the gap, whereas the bleeding domain necessitates approximately 80\%.
Furthermore, we observed a beneficial cross-domain transfer effect: \textbf{integrating auxiliary data from disparate domains enhances global model generalizability}. Notably, supplementing the training set exclusively with data from the smoke domain yielded performance improvements across both the bleeding and low-brightness domains.

\begin{table*}[t]
  \centering
  \caption{Auxiliary training data results. Values are mean $\pm$ standard deviation over three independent training runs differing only in random seed, except the ``uncorrupted'' rows, which are the five-run baselines of Table~\ref{tab:baseline}.}
  \resizebox{0.8\textwidth}{!}{
        \begin{tabular}{c|c|c|c|c|c}
          \toprule
          \multirow{ 2}{*}{\textbf{DSC Score}} & \textbf{Training} & \multirow{ 2}{*}{\textbf{``uncorrupted''}} &  \multirow{ 2}{*}{\textbf{Bleeding}} & \multirow{ 2}{*}{\textbf{Smoke}} & \textbf{Low}  \\
            & \textbf{Domain} & & & &  \textbf{Brightness} \\
            \midrule
            UNet++~\cite{zhou2019unetplusplus} & ``uncorrupted'' 
            & 0.9704 $\pm$ 0.0009 & 0.9306 $\pm$ 0.0371 & 0.8663 $\pm$ 0.0371 & 0.7823 $\pm$ 0.0239 \\   
            UNet++~\cite{zhou2019unetplusplus} & Test 
            & - & 0.9648 $\pm$ 0.0012 & 0.9662 $\pm$ 0.0024 & 0.9400 $\pm$ 0.0061 \\
            UNet++~\cite{zhou2019unetplusplus} & All 
            & 0.9692 $\pm$ 0.0018 & 0.9661 $\pm$ 0.0012 & 0.9689 $\pm$ 0.0017 & 0.9411 $\pm$ 0.0002 \\
            
            Mask2Former~\cite{cheng2021mask2former} & ``uncorrupted''
            & 0.9579 $\pm$ 0.0036 & 0.7501 $\pm$ 0.0280 & 0.8667 $\pm$ 0.0182 & 0.7406 $\pm$ 0.0491 \\
             Mask2Former~\cite{cheng2021mask2former} & Test
             & - & 0.9317 $\pm$ 0.0094 & 0.9466 $\pm$ 0.0087 & 0.8647 $\pm$ 0.0087\\
             Mask2Former~\cite{cheng2021mask2former} & All
            & 0.9556 $\pm$ 0.0022 & 0.9380 $\pm$ 0.0011 & 0.9523 $\pm$ 0.0027 & 0.9104 $\pm$ 0.0050 \\
            \midrule
          \multirow{ 2}{*}{\textbf{NSD Score}} & \textbf{Training} & \multirow{ 2}{*}{\textbf{``uncorrupted''}} & \multirow{ 2}{*}{\textbf{Bleeding}} & \multirow{ 2}{*}{\textbf{Smoke}} & \textbf{Low}  \\
            & \textbf{Domain} & & & &  \textbf{Brightness} \\
            \midrule
            
            UNet++~\cite{zhou2019unetplusplus} & ``uncorrupted''
            & 0.9801 $\pm$  0.0027 & 0.8746 $\pm$ 0.0188 & 0.8357 $\pm$ 0.0374 & 0.6719 $\pm$ 0.0423 \\   
            UNet++~\cite{zhou2019unetplusplus} & Test 
            & - & 0.9660 $\pm$ 0.0056 & 0.9719 $\pm$ 0.0039 & 0.8863 $\pm$ 0.0184 \\
            UNet++~\cite{zhou2019unetplusplus} & All 
            &0.9764 $\pm$ 0.0067& 0.9643 $\pm$ 0.0076 & 0.9774 $\pm$ 0.0076 & 0.8879 $\pm$ 0.0059 \\
            Mask2Former~\cite{cheng2021mask2former} & ``uncorrupted''
            & 0.9523 $\pm$ 0.0100 & 0.6135 $\pm$ 0.0243 & 0.7740 $\pm$ 0.0159 & 0.5866 $\pm$ 0.0415 \\
            Mask2Former~\cite{cheng2021mask2former} & Test
            & -& 0.8905 $\pm$ 0.0144 & 0.9240 $\pm$ 0.0209 & 0.7717 $\pm$ 0.0023\\
            Mask2Former~\cite{cheng2021mask2former} & All
            & 0.9443 $\pm$ 0.0094 & 0.8913 $\pm$ 0.0056 & 0.9379 $\pm$ 0.0107 & 0.8246 $\pm$ 0.0105 \\
          \bottomrule
        \end{tabular}
    }
\label{tab:ood}
\end{table*}

\begin{table*}[t]
  \centering
  \caption{Test results of all participants.}
  \resizebox{\textwidth}{!}{
        \begin{tabular}{c|c|c|c|c|c}
          \toprule
          \textbf{DSC Score} & \textbf{``uncorrupted''} & \textbf{BG Change} & \textbf{Bleeding} & \textbf{Smoke} & \textbf{Low Brightness} \\
          
            \midrule
            SAM2~\cite{ravi2024sam}  
            & 0.9096 & 0.9325 & 0.8628  & 0.8660  & 0.4373 \\
            UNet~\cite{RonnebergerFB15unet} + AutoAugment~\cite{cubuk2019autoaugment} 
            & 0.9568 & 0.9446 & 0.7910 & 0.8895 & 0.6965 \\
            \midrule
            VSI
            & 0.6547 & 0.6161 & 0.6021 & 0.6153 & 0.5805\\
            UBCRCL
            & 0.9587 & \textbf{0.9564} &\textbf{ 0.9424} & 0.9426 & 0.9030\\
            SAM\_KHU
            & 0.8476 & 0.8023 & 0.7644 & 0.7718 & 0.5871\\
            Medical\_Mechatronics
            & 0.6607 & 0.7625 & 0.5079 & 0.5670 & 0.4372\\
            CCG\_DGIST
            & 0.9215 & 0.9008 & 0.7335 & 0.8474 & 0.5796\\
            SK
            & 0.9402 & 0.9493 & 0.9006 & 0.9062 & 0.5365\\
            Tailor
            & 0.9439 & 0.9432 & 0.8902 & 0.9329 & 0.8211\\
            FightTumor
            & 0.9569 & 0.9486 & 0.8528 & 0.9453 & 0.7465\\
            ICVS-2AI
            & \textbf{0.9706} & 0.9530 & 0.9151 & \textbf{0.9657} & \textbf{0.9102} \\
            AIMI
            & 0.9626 & 0.9508 & 0.8517 & 0.8625 & 0.7895\\
            \midrule
          \textbf{NSD Score} & \textbf{``uncorrupted''} & \textbf{BG Change} & \textbf{Bleeding} & \textbf{Smoke} & \textbf{Low Brightness} \\
          
          \midrule
            SAM2~\cite{ravi2024sam}  
            & 0.8479 & 0.9186 & 0.8002  & 0.7808  & 0.2882 \\
            UNet~\cite{RonnebergerFB15unet} + AutoAugment~\cite{cubuk2019autoaugment} 
            & 0.9542 & 0.9196 & 0.6654 & 0.8152 & 0.5344 \\
            \midrule
            VSI
            & 0.5984 & 0.5411 & 0.5023 & 0.5623 & 0.4850 \\
            UBCRCL
            & 0.9699 & \textbf{0.9690} & \textbf{0.9345} & 0.9340 & 0.8268 \\
            SAM\_KHU
            & 0.7531 & 0.6759 & 0.6500 & 0.6540 & 0.4277 \\
            Medical\_Mechatronics
            & 0.3537 & 0.4969 & 0.2627 & 0.3816 & 0.2137 \\
            CCG\_DGIST
            & 0.8523 & 0.7872 & 0.5816 & 0.6882 & 0.4312 \\
            SK
            & 0.8937 & 0.9244 & 0.8099 & 0.8111 & 0.3173 \\
            Tailor
            & 0.9178 & 0.9176 & 0.7884 & 0.8874 & 0.6353 \\
            FightTumor
            & 0.9598 & 0.9279 & 0.7636 & 0.9240 & 0.6116 \\
            ICVS-2AI
            & \textbf{0.9934} & 0.9472 & 0.8870 & \textbf{0.9827} & \textbf{0.8730} \\
            AIMI
            & 0.9693 & 0.9192 & 0.7692 & 0.7763 & 0.6437 \\
          \bottomrule
        \end{tabular}
    }
\label{tab:teams}
\end{table*}

\subsection{Challenge Evaluation}
\label{sec:results_submission}
We evaluated all participant submissions on both the ``uncorrupted'' test domain and the ``corrupted'' test domains, which include background change, bleeding, smoke, and low brightness. Challenge rankings were determined based solely on performance on the latter three domains, as they remained unreleased during the competition to ensure fairness. Among all the teams, Team ICVS-2AI and UBCRCL achieve the strongest overall performance. Team Tailor, FightTumor, SK, and AIMI achieve comparable results to the baseline models. Since we can not perform multiple runs to justify the statistical significance of the challenge results, we only draw inspiration from the results instead of consolidating insight or conclusions. The following paragraphs summarize the key inspirations derived from the evaluation results and the corresponding approaches summarized in Section~\ref{sec:submissions}.

\noindent\textbf{Strong robustness is possible to be achieved without access to ``corrupted'' data when the ``corruption'' is known.} Despite not having access to training data from the ``corrupted'' domains, Team UBCRCL achieved the best results in the background change and bleeding domains, with DSC scores of 0.9564 and 0.9424 and NSD scores of 0.9690 and 0.9345, respectively. In the smoke and low brightness domains, Team ICVS-2AI led with DSC scores of 0.9657 and 0.9102 and NSD scores of 0.9827 and 0.8730, respectively. 

Notably, both winning teams reported customizing their approaches based on the pre-known ``corruption'' types. Thus, we can not assume the robustness they achieved is generalizable to the ``corruptions'' that are not mentioned in this challenge.
In summary, we argue that it is possible to achieve strong robustness when the ``corruption'' type is known, even without the training data from the specific ``corrupted'' domain.

\noindent\textbf{Data augmentation is the most commonly explored method and possibly contributes significantly to achieving the level of robustness of the winning teams.}
8 out of 10 teams mentioned specifically applying data augmentation in their solution, making data augmentation the most commonly explored method.
Besides general data augmentations, some teams leveraged prior knowledge of the ``corruption'' types to implement ``corruption''-specific augmentation strategies. For the smoke ``corruption,'' team UBCRCL, SK, Tailor, FightTumor, and ICVS-2AI apply various specialized smoke augmentations, and all achieve an average of 0.9385 DSC and 0.9078 NSD score. On the contrary, without specialized smoke augmentation, team AIMI, although achieving comparable performance on the ``uncorrupted'' domain, only shows an average of 0.8625 DSC and 0.7763 NSD score. Unlike others that generate fake smoke in rule-based ways, team ICVS-2AI trains a CycleGAN from internal real data for the smoke generation, resulting in the best performance (0.9657 DSC and 0.9827 NSD) in the smoke domain. 

For bleeding ``corruption,'' Team SK used red-colored ellipses to simulate blood, while Team UBCRCL applied general occlusion-based augmentations with a similar masking effect. These strategies yielded the relatively higher performance of 0.9215 DSC and 0.8722 NSD on average. Other teams with comparable ``uncorrupted''-domain performance (Tailor, ICVS-2AI, FightTumor, and AIMI) did not incorporate bleeding-specific augmentation and experienced larger performance deterioration.

In the low-brightness domain, the top performers, UBCRCL and ICVS-2AI, also had the smallest performance degradation. UBCRCL explicitly applied brightness reduction during training, while ICVS-2AI did not use low-brightness-specific augmentation but still performed well, likely due to overall strong model capacity. Even among the top teams, performance deterioration under low brightness remained non-negligible, suggesting that visibility degradation imposes limitations that augmentation alone cannot fully overcome.

\noindent\textbf{Architecture enhancement is another widely attempted direction, but the effectiveness is less straightforward.}
9 out of 10 teams did exploration and a unique architecture choice in their solution. Furthermore, 3 teams (SAM\_KHU, CCG\_DGIST, and AIMI) applied their own specific design of architecture. On one hand, we notice that the winning teams' solution also demonstrates strong performance on the ``uncorrupted'' domain, indicating that the model's robustness is potentially correlated to the model's general learning capacity. On the other hand, the team's focus on the architecture design doesn't generate overall leading performance on the test domains. Thus, the influence of the architecture capacity is not clear in the challenge results.

\subsection{Validation Studies}
In this subsection, we demonstrate a comprehensive analysis of the two primary aspects focused on by the participants. We select UNet++~\cite{zhou2019unetplusplus} and Mask2Former~\cite{cheng2021mask2former} to serve as the representative benchmarks for Convolutional Neural Networks (CNNs) and Transformers, respectively.

\label{sec:results_validation}
\begin{table*}[ht]
  \centering
  \caption{Experiment results of pretraining and augmentation validation. Values are mean $\pm$ standard deviation over five independent runs differing only in random seed. The nnUNet block has no $-$ Pretrain arm because nnUNet trains from scratch by design.}
  \resizebox{\textwidth}{!}{
    \begin{tabular}{c|c|c|c|c|c}
      \toprule
      \multirow{ 2}{*}{\textbf{DSC Score}}  & \multirow{ 2}{*}{\textbf{``uncorrupted''}} & \textbf{Background} & \multirow{ 2}{*}{\textbf{Bleeding}} & \multirow{ 2}{*}{\textbf{Smoke}} & \textbf{Low}  \\
      & & \textbf{Change} & & &  \textbf{Brightness} \\
      \midrule
      UNet++~\cite{zhou2019unetplusplus} 
      & 0.9704 $\pm$ 0.0009 & 0.9457 $\pm$ 0.0224 & 0.9306 $\pm$ 0.0371 & 0.8663 $\pm$ 0.0371 & 0.7823 $\pm$ 0.0239 \\
      - Pretrain
      & 0.9559 $\pm$ 0.0029 & 0.9299 $\pm$ 0.0034 & 0.8134 $\pm$ 0.0214 & 0.8630 $\pm$ 0.0130 & 0.6136 $\pm$ 0.0586 \\
      - Augmentation
      & 0.9146 $\pm$ 0.0179 & 0.8225 $\pm$ 0.0171 & 0.6907 $\pm$ 0.0486 & 0.6784 $\pm$ 0.0190 & 0.5245 $\pm$ 0.0767 \\
      + Customized Augmentation
      & 0.9706 $\pm$ 0.0007 & 0.9464 $\pm$ 0.0327 & 0.9332 $\pm$ 0.0126 & 0.9419 $\pm$ 0.0146 & 0.8496 $\pm$ 0.0267 \\
      \midrule
      Mask2Former~\cite{cheng2021mask2former}
      & 0.9579 $\pm$ 0.0036 & 0.9226 $\pm$ 0.0088 & 0.7501 $\pm$ 0.0280 & 0.8667 $\pm$ 0.0182 & 0.7406 $\pm$ 0.0491 \\
      - Pretrain
      & 0.3084 $\pm$ 0.0000 & 0.3084 $\pm$ 0.0000 & 0.3084 $\pm$ 0.0000 & 0.3084 $\pm$ 0.0000 & 0.3084 $\pm$ 0.0000 \\
      - Augmentation
      & 0.8678 $\pm$ 0.0351 & 0.8142 $\pm$ 0.0130 & 0.5564 $\pm$ 0.0654 & 0.7359 $\pm$ 0.0357 & 0.3450 $\pm$ 0.1460 \\
      + Customized Augmentation
      & 0.9577 $\pm$ 0.0040 & 0.9401 $\pm$ 0.0047 & 0.8400 $\pm$ 0.0103 & 0.9475 $\pm$ 0.0046 & 0.8063 $\pm$ 0.0399 \\
      \midrule
      nnUNet~\cite{isensee2021nnu} Default Augmentation
      & 0.9698 $\pm$ 0.0008 & 0.9555 $\pm$ 0.0054 & 0.7563 $\pm$ 0.0093 & 0.8798 $\pm$ 0.0050 & 0.7251 $\pm$ 0.0569 \\
      AutoAugment
      & 0.9697 $\pm$ 0.0007 & 0.9449 $\pm$ 0.0084 & 0.7521 $\pm$ 0.0063 & 0.8836 $\pm$ 0.0155 & 0.6776 $\pm$ 0.0450 \\
      Customized Augmentation
      & 0.9701 $\pm$ 0.0006 & 0.9552 $\pm$ 0.0085 & 0.7549 $\pm$ 0.0032 & 0.8673 $\pm$ 0.0365 & 0.7136 $\pm$ 0.0452 \\
      No Augmentation
      & 0.9431 $\pm$ 0.0052 & 0.9137 $\pm$ 0.0056 & 0.7264 $\pm$ 0.0065 & 0.8531 $\pm$ 0.0093 & 0.6098 $\pm$ 0.0173 \\
      \midrule
       \multirow{ 2}{*}{\textbf{NSD Score}}  & \multirow{ 2}{*}{\textbf{``uncorrupted''}} & \textbf{Background} & \multirow{ 2}{*}{\textbf{Bleeding}} & \multirow{ 2}{*}{\textbf{Smoke}} & \textbf{Low}  \\
      & & \textbf{Change} & & &  \textbf{Brightness} \\
      \midrule
      UNet++~\cite{zhou2019unetplusplus} 
      & 0.9801 $\pm$ 0.0027 & 0.9351 $\pm$ 0.0362 & 0.8746 $\pm$ 0.0188 & 0.8357 $\pm$ 0.0374 & 0.6719 $\pm$ 0.0423 \\
      - Pretrain
      & 0.9326 $\pm$ 0.0081 & 0.8783 $\pm$ 0.0105 & 0.6561 $\pm$ 0.0212 & 0.7732 $\pm$ 0.0197 & 0.4722 $\pm$ 0.0528 \\
      - Augmentation
      & 0.8320 $\pm$ 0.0182 & 0.6112 $\pm$ 0.0442 & 0.5871 $\pm$ 0.0198 & 0.5225 $\pm$ 0.0107 & 0.4271 $\pm$ 0.0564 \\
      + Customized Augmentation
      & 0.9795 $\pm$ 0.0033 & 0.9430 $\pm$ 0.0385 & 0.8874 $\pm$ 0.0125 & 0.9376 $\pm$ 0.0150 & 0.7535 $\pm$ 0.0311 \\
      \midrule
      Mask2Former~\cite{cheng2021mask2former}
      & 0.9523 $\pm$ 0.0100 & 0.8568 $\pm$ 0.0175 & 0.6135 $\pm$ 0.0243 & 0.7740 $\pm$ 0.0159 & 0.5866 $\pm$ 0.0415 \\
      - Pretrain
      & 0.0955 $\pm$ 0.0000 & 0.0955 $\pm$ 0.0000 & 0.0955 $\pm$ 0.0000 & 0.0955 $\pm$ 0.0000 & 0.0955 $\pm$ 0.0000 \\
      - Augmentation
      & 0.7770 $\pm$ 0.0476 & 0.6450 $\pm$ 0.0095 & 0.4308 $\pm$ 0.0541 & 0.5625 $\pm$ 0.0340 & 0.2308 $\pm$ 0.1235 \\
      + Customized Augmentation
      & 0.9517 $\pm$ 0.0119 & 0.9017 $\pm$ 0.0110 & 0.6930 $\pm$ 0.0144 & 0.9237 $\pm$ 0.0152 & 0.6767 $\pm$ 0.0350 \\
      \midrule
      nnUNet~\cite{isensee2021nnu} Default Augmentation
      & 0.9782 $\pm$ 0.0035 & 0.9343 $\pm$ 0.0182 & 0.6296 $\pm$ 0.0072 & 0.8037 $\pm$ 0.0126 & 0.6053 $\pm$ 0.0432 \\
      AutoAugment
      & 0.9794 $\pm$ 0.0049 & 0.9080 $\pm$ 0.0097 & 0.6174 $\pm$ 0.0119 & 0.8016 $\pm$ 0.0168 & 0.5619 $\pm$ 0.0264 \\
      Customized Augmentation
      & 0.9812 $\pm$ 0.0029 & 0.9368 $\pm$ 0.0223 & 0.6259 $\pm$ 0.0061 & 0.7913 $\pm$ 0.0389 & 0.6000 $\pm$ 0.0433 \\
      No Augmentation
      & 0.8876 $\pm$ 0.0177 & 0.8223 $\pm$ 0.0130 & 0.5588 $\pm$ 0.0084 & 0.7365 $\pm$ 0.0142 & 0.4972 $\pm$ 0.0220 \\
      \bottomrule
    \end{tabular}
    }
\label{tab:augmentation}
\end{table*}

\subsubsection{Augmentation Validation}
\begin{figure*}[ht]
\centering
\includegraphics[width=\linewidth]{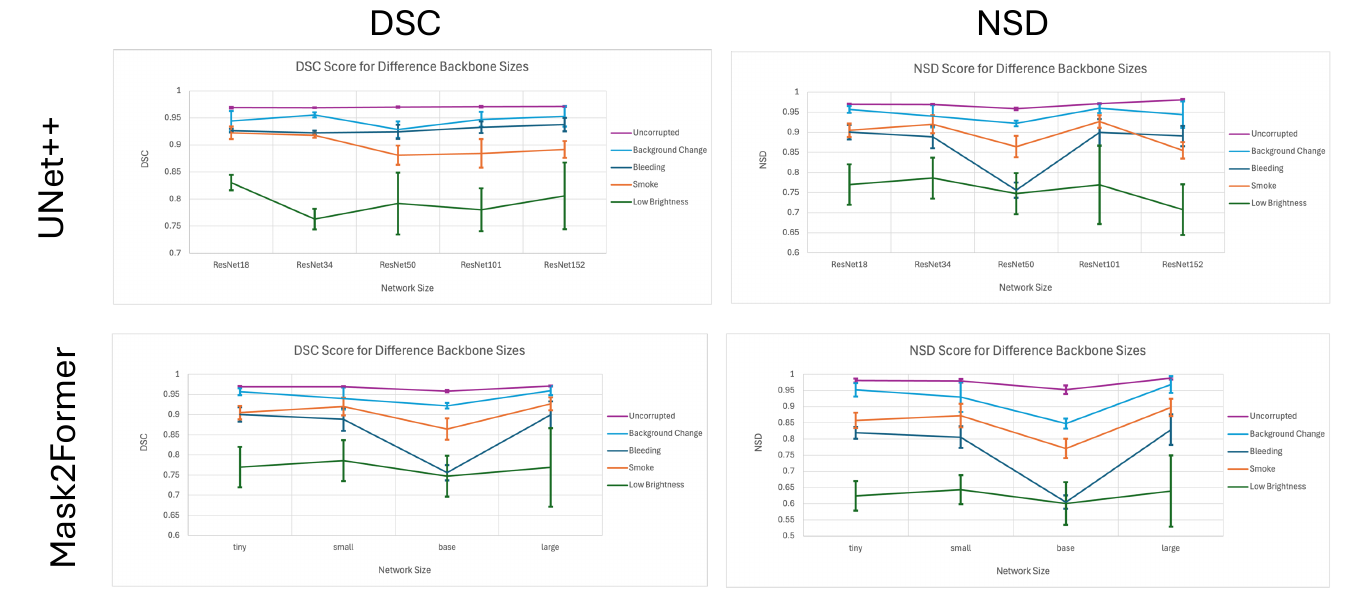}
\caption{Validation study results on network sizes.}
\label{fig:network_size}
\end{figure*}

In this study, we empirically validate the critical role of data augmentation in model robustness. Our analysis centers on the two primary categories most prevalent in the field: (1) General Data Augmentation and (2) Customized Data Augmentation.
As detailed in Table~\ref{tab:augmentation}, ablating general augmentation from the baseline precipitates a sharp performance decline across all ``corrupted'' test domains, confirming that \textbf{general data augmentation plays a foundational role in maintaining robustness}. To evaluate the impact of customized augmentation, we synthesized strategies from challenge participants to formulate a customized augmentation, as detailed in Section~\ref{sec:baseline}. Quantitative results demonstrate that superimposing this customized augmentation onto standard augmentations yields distinct performance gains across all ``corrupted'' domains, with particularly pronounced efficacy in the smoke and low-brightness scenarios. These findings indicate that \textbf{when potential environmental ``corruptions'' can be anticipated, task-specific data augmentation serves as a potent mechanism for enhancing model robustness.} Table~\ref{tab:augmentation} also reports nnUNet run unmodified, which qualifies this. Against its own self-configured pipeline, neither challenge-specific scheme helps: the customized augmentation differs from its default augmentation in no domain ($p \geq 0.45$), and AutoAugment is $0.0145$ DSC lower on the ``corrupted'' average. Disabling augmentation entirely costs $0.0535$. For a framework whose augmentation is already tuned to the data, therefore, \textbf{the benefit attributed to augmentation comes from having a mature pipeline rather than from tailoring one to the anticipated ``corruption'' types.}

\subsubsection{Network Architecture Validation}
In this study, we empirically investigate the impact of architectural decisions on model robustness, focusing on two foundational factors: (1) Pretrained Initialization and (2) Model Capacity.
As detailed in Table~\ref{tab:augmentation}, the absence of pretraining leads to significant degradation. UNet++ exhibits a remarkable performance drop in the bleeding and low-brightness domains, while Mask2Former suffers from severe optimization collapse, resulting in it being untrainable from scratch. \textbf{These findings underscore the indispensability of pretrained weights for stabilizing convergence and ensuring generalization.} 
To assess the influence of model capacity, we evaluated UNet++ across a spectrum of backbones (ResNet18 through ResNet152~\cite{he2016deep}) and Mask2Former across varying Swin Transformer sizes (Tiny to Large~\cite{liu2021swin}).
Contrary to the pretraining results, the data in Figure~\ref{fig:network_size} indicates that ~\textbf{increasing network depth yields negligible benefits in this context, suggesting that model capacity is not the primary bottleneck for robustness.} 

This suggests that the disparities in robustness observed across different models are primarily attributable to the inductive biases inherent in their architectural designs. However, given the significant lack of interpretability in these systems, purposefully engineering such biases to enhance robustness remains a formidable challenge.

\subsubsection{Robustness to Undisclosed ``corruptions'' and to a Combination}

The studies above evaluate ``corruption'' types known in advance. To ask which ingredient confers robustness on unanticipated ones, we repeated the ablation of Table~\ref{tab:augmentation} on four types never disclosed during the challenge --- defocus blur, specular overexposure, colour shift and compression artefacts, at three severities each --- plus a combination of two disclosed types, smoke with reduced brightness. This is inference only, over the same five runs per arm. The four withheld types are synthetic, the construction we criticise in Section~\ref{sec:intro}, so we read the study as a generalization probe over our own baselines rather than as evidence about real surgery.

\begin{table*}[t]
  \centering
  \caption{DSC on ``corruption'' types withheld from the challenge, UNet++. The arms are those of Table~\ref{tab:augmentation}; values are mean $\pm$ standard deviation over five independent training runs, evaluated without retraining. The four withheld types are synthetic and the last row is physically acquired smoke with synthetic darkening, so this table probes generalization over our own baselines rather than robustness in real surgery.}
  \label{tab:heldout_unetpp}
  \resizebox{0.92\textwidth}{!}{
    \begin{tabular}{l|cccc}
      \toprule
      \textbf{Condition} & \textbf{UNet++} & \textbf{$-$ Pretrain} & \textbf{$-$ Augmentation} & \textbf{$+$ Cust.\ Aug.} \\
      \midrule
      ``Uncorrupted'' reference & 0.9704 $\pm$ 0.0009 & 0.9576 $\pm$ 0.0024 & 0.9146 $\pm$ 0.0170 & 0.9720 $\pm$ 0.0017 \\
      \midrule
      Defocus blur, mild & 0.9689 $\pm$ 0.0011 & 0.9541 $\pm$ 0.0046 & 0.9059 $\pm$ 0.0202 & 0.9626 $\pm$ 0.0073 \\
      \quad moderate & 0.9426 $\pm$ 0.0123 & 0.9428 $\pm$ 0.0055 & 0.8542 $\pm$ 0.0381 & 0.8941 $\pm$ 0.0403 \\
      \quad strong & 0.8749 $\pm$ 0.0256 & 0.9233 $\pm$ 0.0088 & 0.7582 $\pm$ 0.0444 & 0.7939 $\pm$ 0.0677 \\
      \midrule
      Specular overexposure, mild & 0.9609 $\pm$ 0.0022 & 0.9444 $\pm$ 0.0038 & 0.8606 $\pm$ 0.0187 & 0.9666 $\pm$ 0.0012 \\
      \quad moderate & 0.9440 $\pm$ 0.0029 & 0.9297 $\pm$ 0.0033 & 0.7861 $\pm$ 0.0311 & 0.9552 $\pm$ 0.0034 \\
      \quad strong & 0.8971 $\pm$ 0.0098 & 0.8816 $\pm$ 0.0066 & 0.6656 $\pm$ 0.0399 & 0.9124 $\pm$ 0.0203 \\
      \midrule
      Colour shift, mild & 0.9707 $\pm$ 0.0010 & 0.9558 $\pm$ 0.0023 & 0.9120 $\pm$ 0.0200 & 0.9717 $\pm$ 0.0015 \\
      \quad moderate & 0.9701 $\pm$ 0.0013 & 0.9173 $\pm$ 0.0235 & 0.9017 $\pm$ 0.0265 & 0.9706 $\pm$ 0.0013 \\
      \quad strong & 0.9687 $\pm$ 0.0017 & 0.8097 $\pm$ 0.0629 & 0.8874 $\pm$ 0.0342 & 0.9687 $\pm$ 0.0012 \\
      \midrule
      Compression, mild & 0.9647 $\pm$ 0.0011 & 0.9443 $\pm$ 0.0031 & 0.8952 $\pm$ 0.0138 & 0.9604 $\pm$ 0.0018 \\
      \quad moderate & 0.9588 $\pm$ 0.0018 & 0.9383 $\pm$ 0.0037 & 0.8848 $\pm$ 0.0083 & 0.9519 $\pm$ 0.0066 \\
      \quad strong & 0.8815 $\pm$ 0.0144 & 0.8356 $\pm$ 0.0114 & 0.7700 $\pm$ 0.0116 & 0.8669 $\pm$ 0.0313 \\
      \midrule
      Smoke $+$ low brightness & 0.8938 $\pm$ 0.0244 & 0.8838 $\pm$ 0.0099 & 0.6491 $\pm$ 0.0386 & 0.9515 $\pm$ 0.0062 \\
      \bottomrule
    \end{tabular}
  }
\end{table*}

\begin{table*}[t]
  \centering
  \caption{DSC on ``corruption'' types withheld from the challenge, Mask2Former; conventions as in Table~\ref{tab:heldout_unetpp}. Without COCO initialization Mask2Former collapses to a single mask that does not depend on the input (Table~\ref{tab:augmentation}). One ground truth is shared across domains, so that arm scores $0.3084$ on every condition: its flat profile reflects a model that does not read the image, not robustness.}
  \label{tab:heldout_m2f}
  \resizebox{0.92\textwidth}{!}{
    \begin{tabular}{l|cccc}
      \toprule
      \textbf{Condition} & \textbf{Mask2Former} & \textbf{$-$ Pretrain} & \textbf{$-$ Augmentation} & \textbf{$+$ Cust.\ Aug.} \\
      \midrule
      ``Uncorrupted'' reference & 0.9579 $\pm$ 0.0036 & 0.3084 $\pm$ 0.0000 & 0.8679 $\pm$ 0.0351 & 0.9625 $\pm$ 0.0082 \\
      \midrule
      Defocus blur, mild & 0.9564 $\pm$ 0.0037 & 0.3084 $\pm$ 0.0000 & 0.8752 $\pm$ 0.0326 & 0.9576 $\pm$ 0.0078 \\
      \quad moderate & 0.9518 $\pm$ 0.0036 & 0.3084 $\pm$ 0.0000 & 0.8743 $\pm$ 0.0315 & 0.9469 $\pm$ 0.0068 \\
      \quad strong & 0.9437 $\pm$ 0.0034 & 0.3084 $\pm$ 0.0000 & 0.8666 $\pm$ 0.0310 & 0.9261 $\pm$ 0.0060 \\
      \midrule
      Specular overexposure, mild & 0.9457 $\pm$ 0.0052 & 0.3084 $\pm$ 0.0000 & 0.8460 $\pm$ 0.0300 & 0.9483 $\pm$ 0.0156 \\
      \quad moderate & 0.9264 $\pm$ 0.0087 & 0.3084 $\pm$ 0.0000 & 0.7783 $\pm$ 0.0266 & 0.9332 $\pm$ 0.0182 \\
      \quad strong & 0.8690 $\pm$ 0.0142 & 0.3084 $\pm$ 0.0000 & 0.6697 $\pm$ 0.0186 & 0.8985 $\pm$ 0.0160 \\
      \midrule
      Colour shift, mild & 0.9571 $\pm$ 0.0027 & 0.3084 $\pm$ 0.0000 & 0.8318 $\pm$ 0.0360 & 0.9589 $\pm$ 0.0096 \\
      \quad moderate & 0.9430 $\pm$ 0.0058 & 0.3084 $\pm$ 0.0000 & 0.7010 $\pm$ 0.0305 & 0.9433 $\pm$ 0.0159 \\
      \quad strong & 0.9021 $\pm$ 0.0226 & 0.3084 $\pm$ 0.0000 & 0.6048 $\pm$ 0.0254 & 0.9118 $\pm$ 0.0281 \\
      \midrule
      Compression, mild & 0.9391 $\pm$ 0.0063 & 0.3084 $\pm$ 0.0000 & 0.8328 $\pm$ 0.0414 & 0.9416 $\pm$ 0.0084 \\
      \quad moderate & 0.9151 $\pm$ 0.0090 & 0.3084 $\pm$ 0.0000 & 0.8246 $\pm$ 0.0364 & 0.9242 $\pm$ 0.0085 \\
      \quad strong & 0.8138 $\pm$ 0.0069 & 0.3084 $\pm$ 0.0000 & 0.7363 $\pm$ 0.0226 & 0.8195 $\pm$ 0.0183 \\
      \midrule
      Smoke $+$ low brightness & 0.9128 $\pm$ 0.0107 & 0.3084 $\pm$ 0.0000 & 0.6581 $\pm$ 0.0666 & 0.9515 $\pm$ 0.0084 \\
      \bottomrule
    \end{tabular}
  }
\end{table*}

Tables~\ref{tab:heldout_unetpp} and~\ref{tab:heldout_m2f} show that \textbf{general augmentation, not the targeted kind, is what transfers.} Averaged over the twelve withheld conditions, removing it costs $0.102$ DSC for UNet++ and $0.135$ for Mask2Former --- roughly twice its cost on ``uncorrupted'' data --- and it is the only ablation to lower retention materially ($0.971$ to $0.919$, $0.963$ to $0.907$). The customized augmentation is near-neutral outside the ``corruptions'' it simulates ($-0.011$ and $+0.004$), its one systematic cost falling under defocus blur, which is spatial where its heuristics are photometric; on the combination of two anticipated types it gains $0.058$ and $0.039$. Targeted augmentation is therefore worth what it was designed for, and general augmentation is what buys robustness beyond it.

\subsubsection{Insight Generalization}
To assess the external validity of our findings, we extended our analysis to independent surgical datasets and tasks. We selected CholecTrack20~\cite{nwoye2025cholectrack20} as the primary evaluation benchmark, as it inherently contains naturally occurring instances of bleeding and smoke. Adopting YOLOv11~\cite{khanam2024yolov11} as the baseline architecture, we employed an ablation protocol analogous to our previous experiments to isolate the contributions of data augmentation and pretraining, reporting standard mAP50 and mAP50-95 as the primary metrics.
As detailed in Table~\ref{tab:generalization}, the baseline model demonstrates intrinsic robustness against smoke and bleeding, a result attributable to the inclusion of these artifacts within the training distribution. However, the ablation of general data augmentation and pretraining precipitates a marked performance degradation across all test domains, with the degradation being notably more pronounced in 'corrupted' domains compared to the aggregate test set. While the introduction of customized augmentation for bleeding and smoke yields overall performance improvements, the gains in certain scenarios are not significant likely due to the pre-existing representation of these ``corruption'' types in the training data. In summary, these findings align consistently with the SegSTRONG-C results, effectively corroborating the generalizability of our derived insights.

\begin{table*}[ht]
  \centering
  \caption{Experiment results of YOLOv11 on Cholectrack20~\cite{nwoye2025cholectrack20}. Values are mean $\pm$ standard deviation over five independent runs.}
  \resizebox{\textwidth}{!}{
    \begin{tabular}{c|c|c|c|c}
      \toprule
      \textbf{mAP50}  & \textbf{All test} & \textbf{Bleeding only} & \textbf{Smoke Only} &\textbf{Bleeding \& Smoke}\\
      \midrule
      YOLOv11~\cite{khanam2024yolov11} 
      & 0.7610 $\pm$ 0.0075 & 0.7358 $\pm$ 0.0131 & 0.7381 $\pm$ 0.0170 & 0.7381 $\pm$ 0.0219 \\
      - Pretrain
      & 0.7369 $\pm$ 0.0132 & 0.7159 $\pm$ 0.0164 & 0.7125 $\pm$ 0.0181 & 0.7006 $\pm$ 0.0141 \\
      - Augmentation
      & 0.6147 $\pm$ 0.0102 & 0.5883 $\pm$ 0.0077 & 0.5530 $\pm$ 0.0434 & 0.4971 $\pm$ 0.0345 \\
      + Customized Augmentation
      & 0.7778 $\pm$ 0.0089  &  0.7578  $\pm$ 0.0087 &  0.7399 $\pm$ 0.0141 & 0.7461 $\pm$ 0.0115
      \\
      \midrule
      \textbf{mAP50-95}  & \textbf{All test} & \textbf{Bleeding only} & \textbf{Smoke Only} &\textbf{Bleeding \& Smoke}\\
      \midrule
      YOLOv11~\cite{khanam2024yolov11} 
      & 0.5267 $\pm$ 0.0025 & 0.5010 $\pm$ 0.0081 & 0.4944 $\pm$ 0.0182 & 0.4817 $\pm$ 0.0244  \\
      - Pretrain
       & 0.4958 $\pm$ 0.0098 & 0.4744 $\pm$ 0.0117 & 0.4544 $\pm$ 0.0140 & 0.4342 $\pm$ 0.0141 \\
      - Augmentation
      & 0.4000 $\pm$ 0.0056 & 0.3810 $\pm$ 0.0029 & 0.3597 $\pm$ 0.0358 & 0.3186 $\pm$ 0.0293 \\
      + Customized Augmentation
      & 0.5362 $\pm$ 0.0048  & 0.5132 $\pm$ 0.0053 & 0.4969 $\pm$ 0.0178 & 0.4866 $\pm$ 0.0214 \\
      \bottomrule
    \end{tabular}
    }
\label{tab:generalization}
\end{table*}

\subsection{Extended Discussion}
While benchmark performance remains a standard metric for algorithm development, it alone is insufficient to evaluate model reliability under the variable conditions of real-world surgery. As our baseline analysis confirmed, ``naively trained'' end-to-end DNNs that excel in ``uncorrupted'' domains often suffer substantial degradation when exposed to non-adversarial ``corruptions'' such as smoke, bleeding, low lighting, and background variation. In the context of high-stakes surgical applications, overlooking this robustness concern during development can result in severe clinical consequences. SegSTRONG-C is designed to investigate these scenarios for robotic tool segmentation and provide a benchmark for evaluating algorithmic robustness. By using a robotic trajectory replay and a reproducible data generation pipeline, "corrupted" test data can be readily synthesized with high-quality annotations for evaluation.

In our baseline study, we quantitatively demonstrate the \textbf{performance degradation in both end-to-end trained neural networks and zero-shot foundation models}. To elucidate the mechanisms of this decline, we disentangled two contributing factors: domain shift and reduced visibility. Through customized training on ``corrupted'' data, we confirm that \textbf{domain shift plays a major role in performance deterioration}. However, the persistence of performance gaps even under domain-aligned training highlights that \textbf{it is crucial to enhance models’ ability to extract meaningful features from visually degraded inputs}. Furthermore, our analysis revealed that \textbf{while incorporating ``corrupted'' data can mitigate domain shift and improve global generalization, doing so requires significant data collection efforts (ranging from 20\% to 80\% of the ``uncorrupted'' training set), which is often infeasible for clinical datasets.}  We emphasize that the zero-``corrupted''-data condition is the challenge setting rather than a recommendation for dataset development. Where real ``corrupted'' examples exist, our results in Table~\ref{tab:ood} indicate that they should be used: mixed-domain training over all available domains recovers all of the lost performance essentially, the required volume is ``corruption''-dependent, and auxiliary data from one domain transfers to others. The practical recommendation is therefore to maximize inclusion of whatever real ``corrupted'' data is available, and to treat synthetic augmentation as the fallback when it is not.

\begin{table*}[t]
  \centering
  \caption{Customized augmentation added on top of real ``corrupted'' training data. Training on all ``uncorrupted'' data plus auxiliary ``corrupted'' data at three ratios of the ``uncorrupted'' set, each with and without the customized augmentation. One run per cell, trained in a single environment, so the comparison is within each pair rather than against Table~\ref{tab:ood}.}
  \label{tab:aug_on_top}
  \resizebox{\textwidth}{!}{
    \begin{tabular}{c|c|c|c|c|c}
      \toprule
      \multirow{2}{*}{\textbf{DSC Score}} & \multirow{2}{*}{\textbf{``uncorrupted''}} & \textbf{Background} & \multirow{2}{*}{\textbf{Bleeding}} & \multirow{2}{*}{\textbf{Smoke}} & \textbf{Low} \\
      & & \textbf{Change} & & & \textbf{Brightness} \\
      \midrule
      ``Uncorrupted'' $+$ 20\% auxiliary data
      & 0.9639 & 0.9514 & 0.9112 & 0.9569 & 0.7831 \\
      \quad $+$ Customized Augmentation
      & 0.9697 & 0.9646 & 0.9520 & 0.9634 & 0.8834 \\
      \midrule
      ``Uncorrupted'' $+$ 60\% auxiliary data
      & 0.9630 & 0.8715 & 0.9227 & 0.9576 & 0.9046 \\
      \quad $+$ Customized Augmentation
      & 0.9512 & 0.9151 & 0.9317 & 0.9467 & 0.9022 \\
      \midrule
      ``Uncorrupted'' $+$ 100\% auxiliary data
      & 0.9631 & 0.8767 & 0.9334 & 0.9649 & 0.8739 \\
      \quad $+$ Customized Augmentation
      & 0.9057 & 0.8296 & 0.9088 & 0.9051 & 0.8892 \\
      \midrule
      \multirow{2}{*}{\textbf{NSD Score}} & \multirow{2}{*}{\textbf{``uncorrupted''}} & \textbf{Background} & \multirow{2}{*}{\textbf{Bleeding}} & \multirow{2}{*}{\textbf{Smoke}} & \textbf{Low} \\
      & & \textbf{Change} & & & \textbf{Brightness} \\
      \midrule
      ``Uncorrupted'' $+$ 20\% auxiliary data
      & 0.9595 & 0.9378 & 0.8325 & 0.9441 & 0.6886 \\
      \quad $+$ Customized Augmentation
      & 0.9807 & 0.9656 & 0.9207 & 0.9690 & 0.7883 \\
      \midrule
      ``Uncorrupted'' $+$ 60\% auxiliary data
      & 0.9606 & 0.8633 & 0.8671 & 0.9531 & 0.8255 \\
      \quad $+$ Customized Augmentation
      & 0.9469 & 0.8975 & 0.8921 & 0.9387 & 0.8029 \\
      \midrule
      ``Uncorrupted'' $+$ 100\% auxiliary data
      & 0.9638 & 0.8600 & 0.9050 & 0.9683 & 0.7782 \\
      \quad $+$ Customized Augmentation
      & 0.8699 & 0.8084 & 0.8577 & 0.8662 & 0.7811 \\
      \bottomrule
    \end{tabular}
  }
\end{table*}

Table~\ref{tab:aug_on_top} tests directly whether synthetic augmentation also adds value on top of real ``corrupted'' data, by adding the customized augmentation to training on all ``uncorrupted'' data plus 20\%, 60\% and 100\% auxiliary ``corrupted'' data, and comparing against a matched baseline at each ratio. It does not, uniformly: the gain is $+0.0402$ DSC on the ``corrupted'' average at 20\%, falls to $+0.0098$ at 60\%, and becomes $-0.0290$ at 100\%, where the customized augmentation also costs performance on the ``uncorrupted'' domain. Synthetic augmentation therefore substitutes for real ``corrupted'' data rather than supplementing it, and once the real data is available it can over-augment; we report this comparison at a single seed per ratio and so treat the crossover as approximate.

Through the SegSTRONG-C challenge, we observed that top-performing teams achieved robust performance without access to ``corrupted''-domain data, primarily by leveraging prior knowledge of ``corruption'' types for data augmentation and selecting suitable architectures. Inspired by these approaches, we conducted validation studies to empirically consolidate these insights via comprehensive controlled experiments. As a practical method, \textbf{general data augmentation offers a cost-effective and annotation-free approach to improve feature extraction.} Furthermore, \textbf{customized data augmentation emerged as the most common and effective strategy when the ``corruption'' type can be anticipated,} enabling targeted mitigation of domain shift. To this end, we provide a straightforward, customized data augmentation pipeline to address potential bleeding, smoke, and low-brightness scenarios. Beyond augmentation, our studies also explored the impact of architectural choices. \textbf{As a foundational strategy, loading pretrained weights is proven to be effective and essential in facilitating convergence and enhancing model robustness.} Conversely, \textbf{simply increasing network capacity does not guarantee improvement in model robustness.}

In summary, to develop a robust model for surgical applications, we recommend the following guidelines that is feasible to apply: (1) assess the availability of diverse data sources and maximize the inclusion of training data; (2) utilize pretrained weights for initialization; (3) apply general data augmentation during training; and (4) customize synthetic data augmentation to target ``corruption'' types that are anticipated yet scarce in the original training set.

\subsection{Limitations}
Despite our achievement, the SegSTRONG-C challenge leaves several open questions. All ``corruption'' types were disclosed in advance, allowing teams to engineer targeted solutions. However, in real surgical environments, the nature and severity of visual ``corruptions'' are not always known. As such, the robustness of current submissions to unknown or unseen non-adversarial ``corruptions'' remains less explored. Team UBCRCL, however, not only applied customized approaches but also attempted a geometry-aware method. They incorporated pseudo-depth generated from foundation models (e.g., DepthAnything) to form RGB-D feature encodings, yielding top-tier performance on several ``corruption'' types. This suggests that geometric representations may provide a more invariant, unified view of the scene, which generalizes more effectively under ``corruption.'' The rise of foundation models capable of extracting such geometric priors and strong zero-shot generalization makes this direction inspiring. However, the current attempts around foundation models (e.g., finetuning on limited surgical data) do not provide the expected outcome. Other ongoing research efforts aiming for more generalizable feature extraction, such as domain generalization~\cite{seenivasan2022biomimetic,reiter2023domain,philipp2022dynamic}, also merit validation. 

SegSTRONG-C should also be read as a controlled, diagnostic benchmark rather than a comprehensive proxy for robustness in real surgery. Its scenario diversity is deliberately limited in several respects. The dataset comprises 17 sequences acquired in a single laboratory from seven camera and robot configurations, with ex-vivo animal tissue standing in for the operative field, and the held-out test set is drawn from a single configuration. Trajectories are teleoperated in free space and do not include tool--tissue interaction, so the tool poses are independent of the ``corruption'' -- this is the property that makes the counterfactual comparison clean, but it also means the scenes lack the manipulation, deformation, and occlusion that characterize real procedures. Annotations are binary tool masks, without instance or part labels. The ``corruptions'' are four physically induced conditions at a single severity each, chosen for reproducibility under kinematic replay, and they do not span the full range of visual degradation encountered intraoperatively. Consequently, results on SegSTRONG-C should be interpreted as evidence about the mechanisms of model failure under controlled appearance shift and visibility loss, and as a comparative diagnostic across methods -- not as a prediction of clinical performance.

\section{Conclusion}
We presented SegSTRONG-C, a comprehensive benchmark and community challenge dedicated to evaluating algorithmic robustness against realistic, non-adversarial ``corruptions'' in surgical data science. By systematically simulating the diverse visual ``corruptions'' inherent to the intraoperative environment, SegSTRONG-C addresses a critical deficiency in current evaluation methodologies, establishing an essential framework for validating deep learning systems under conditions that closely mirror the complexity of the real-world operating theater.

The empirical results from SegSTRONG-C offer a rigorous assessment of contemporary robustness strategies. Our findings demonstrate that while state-of-the-art baseline models suffer marked degradation under ``corruption,'' performance can be substantially recovered, even without access to ``corrupted'' target data, through the strategic application of customized data augmentation. These insights provide clear, actionable guidance for the engineering of future surgical AI systems. By delineating effective practices and exposing the limitations of existing paradigms, SegSTRONG-C lays the foundation for the advancement of clinically dependable and ``corruption''-aware surgical intelligence.

\bibliographystyle{IEEEtran}
\bibliography{main}

\end{document}